\documentclass[11pt]{article}
\usepackage[utf8]{inputenc}
\usepackage[T1]{fontenc}
\usepackage{lmodern}
\usepackage[margin=1in]{geometry}
\usepackage{microtype}
\usepackage{amsmath,amssymb}
\usepackage{graphicx}
\usepackage{booktabs,longtable,array,tabularx}
\usepackage{xcolor}
\usepackage{enumitem}
\usepackage{caption}
\usepackage{rotating}
\usepackage{tikz}
\usetikzlibrary{positioning,arrows.meta,fit,calc,patterns}
\usepackage[numbers,sort&compress]{natbib}
\usepackage[colorlinks=true,linkcolor=blue!50!black,citecolor=blue!50!black,urlcolor=blue!50!black]{hyperref}

\newcommand{\npr}{\textsuperscript{$\ast$}}
\newcolumntype{L}[1]{>{\raggedright\arraybackslash}p{#1}}
\setlist{itemsep=2pt,topsep=4pt}
\title{Vision-Language-Action Models:\\
A Life-Cycle Survey of Design, Training and Deployment}
\author{Suvarna Kadam\\
\small Witsense AI\\
\small \texttt{suvarna@witsense.ai}
\and
Soran Parsa\\
\small Department of Computer Science, University of Huddersfield\\
\small \texttt{s.parsa@hud.ac.uk}}
\date{November 2026}

\begin{document}
\maketitle

\begin{abstract}
Vision-language-action (VLA) models adapt web-pretrained vision-language models to output robot actions, so that web knowledge carries over to physical control. In under four years the field has grown from RT-2, the first VLA, to open models that fine-tune on a single consumer GPU and proprietary generalists that clean unseen homes. VLA research joins robotics, computer vision and language modelling, so this survey is written for researchers and practitioners from each of these fields. It reviews 89 works from December 2022 to October 2026, mainly on manipulation and navigation, plus humanoid control, world-action models and evaluation. Seven design axes follow a VLA's life cycle: model design (inputs and memory, architecture, action output, intermediate predictions), training (pretraining data, post-training) and deployment (real-time execution). We compare 22 models on these axes, with their training data, model size, inference speed and hardware. Three trends emerge. (i)~Training data has broadened beyond teleoperated robot data (about 10,000 hours for $\pi_0$) to community-collected episodes and egocentric human video (20,000 hours for GR00T~N1.7). (ii)~Model sizes have split into two groups: most recent models have 3--7 billion parameters; a new group under one billion runs on consumer GPUs. (iii)~Generalist VLAs handling both manipulation and navigation are emerging, such as UniVLA and Qwen-VLA. We also compare VLAs with world-action models, which predict future observations as well as actions, and review evidence that some VLAs with near-perfect benchmark scores fail after small scene changes. A practitioner's guide gives a default for each design decision, and a glossary supports readers new to the field.
\end{abstract}

\section{Introduction}

Large language and vision-language models (VLMs) acquire broad semantic knowledge from internet-scale data. Robot learning, by contrast, has historically trained a separate policy for each robot, task and environment, using datasets several orders of magnitude smaller. Vision-language-action (VLA) models connect the two. A VLA starts from a pretrained VLM and fine-tunes it to emit actions alongside, or instead of, text. The term was introduced with RT-2~\citep{rt2}, which expressed robot actions as text tokens, co-fine-tuned a VLM on robot trajectories and web data, and showed emergent semantic generalisation. Since then the design space has grown at every stage of a VLA's life cycle. In the model, action heads now range from discrete tokens to diffusion and flow matching, and pipelines from single networks to dual-system and hierarchical designs. In training, data has grown from tens of thousands of teleoperated episodes to tens of thousands of hours of multi-robot experience, supplemented by human video, simulation and the robot's own autonomous attempts. In deployment, models that compute at a few hertz must still control robots smoothly at tens of hertz.

This survey has five aims:
\begin{itemize}
  \item \textbf{A life-cycle taxonomy.} We organise VLA research along seven design axes that follow a VLA's life cycle, from model design through training to deployment (Section~\ref{sec:taxonomy}), so that a new model can be placed by the choices it makes on each; Table~\ref{tab:axesgrid} places 22 main models this way.
  \item \textbf{Manipulation and navigation side by side.} Most VLA surveys concentrate on manipulation. We treat navigation VLAs as a parallel lineage (Section~\ref{sec:navigation}), cover locomotion and humanoid whole-body control (Section~\ref{sec:humanoid}), and discuss the recent convergence of manipulation and navigation (Section~\ref{sec:convergence}).
  \item \textbf{A data-centred quantitative view.} For every work we extract the training data each paper reports and plot the field's trajectory in robot episodes and in parameter count (Figure~\ref{fig:scaling}, Tables~\ref{tab:manip} and~\ref{tab:nav}).
  \item \textbf{VLAs versus world-action models, and fair evaluation.} We compare VLAs with world-action models (Section~\ref{sec:wam}) and review standard benchmarks and the recent efforts to make VLA evaluation fair and predictive of real-world performance (Section~\ref{sec:benchmarks}).
  \item \textbf{Open problems}, including a reporting problem we encountered directly: training data is described in incompatible units, or not quantified at all, in a large share of papers (Section~\ref{sec:challenges}).
\end{itemize}

\begin{figure}[!t]
\centering
\fbox{\begin{minipage}{0.96\linewidth}
\footnotesize
\setlist{itemsep=0pt,topsep=2pt}
\textbf{At a glance.} The seven design axes (Section~\ref{sec:taxonomy}, Figure~\ref{fig:lifecycle}) follow a VLA through its life cycle.
\begin{itemize}
  \item \textit{Model: what it is.}
  \begin{itemize}
    \item \textbf{Inputs and memory}: what the model observes (cameras, language, robot state, extra conditioning) and how much of the past it keeps.
    \item \textbf{Architecture}: the structure of the whole pipeline, not the backbone network: one network, a dual system of a slow VLM and a fast action module, or a hierarchy of separately trained models.
    \item \textbf{Action output}: how it turns vision-language features into motor commands, from discrete tokens to diffusion, flow matching and regression, usually as a chunk of future actions.
    \item \textbf{Intermediate predictions}: what it predicts besides actions, either reasoning in text (chain of thought, subtasks) or future observations from a world model, which adds a training signal about physical dynamics.
  \end{itemize}
  \item \textit{Training: how it learns.}
  \begin{itemize}
    \item \textbf{Pretraining data}: what it learns from (robot demonstrations, data pooled across robot bodies, web data, human video, simulation and generated trajectories) and how these are combined.
    \item \textbf{Post-training}: how it is adapted and improved for a target robot or task, from fine-tuning recipes to reinforcement learning from deployment.
  \end{itemize}
  \item \textit{Deployment: how it runs.}
  \begin{itemize}
    \item \textbf{Real-time execution}: how a model that runs at a few hertz still controls a robot smoothly, through action chunking, asynchronous inference and splits between a remote server and the robot.
  \end{itemize}
\end{itemize}
Three trends emerge:
\begin{itemize}
  \item \textbf{Broader training data}: from about 10k hours of teleoperated robot data ($\pi_0$), to 23k community-collected episodes (SmolVLA), to large-scale egocentric human video (20k hours for GR00T~N1.7);
  \item \textbf{Model sizes have split}: most recent models have 3--7B parameters, while a new group under one billion runs on consumer GPUs;
  \item \textbf{Generalist VLAs}: models that handle both manipulation and navigation are emerging (UniVLA, Qwen-VLA).
\end{itemize}
\end{minipage}}
\end{figure}

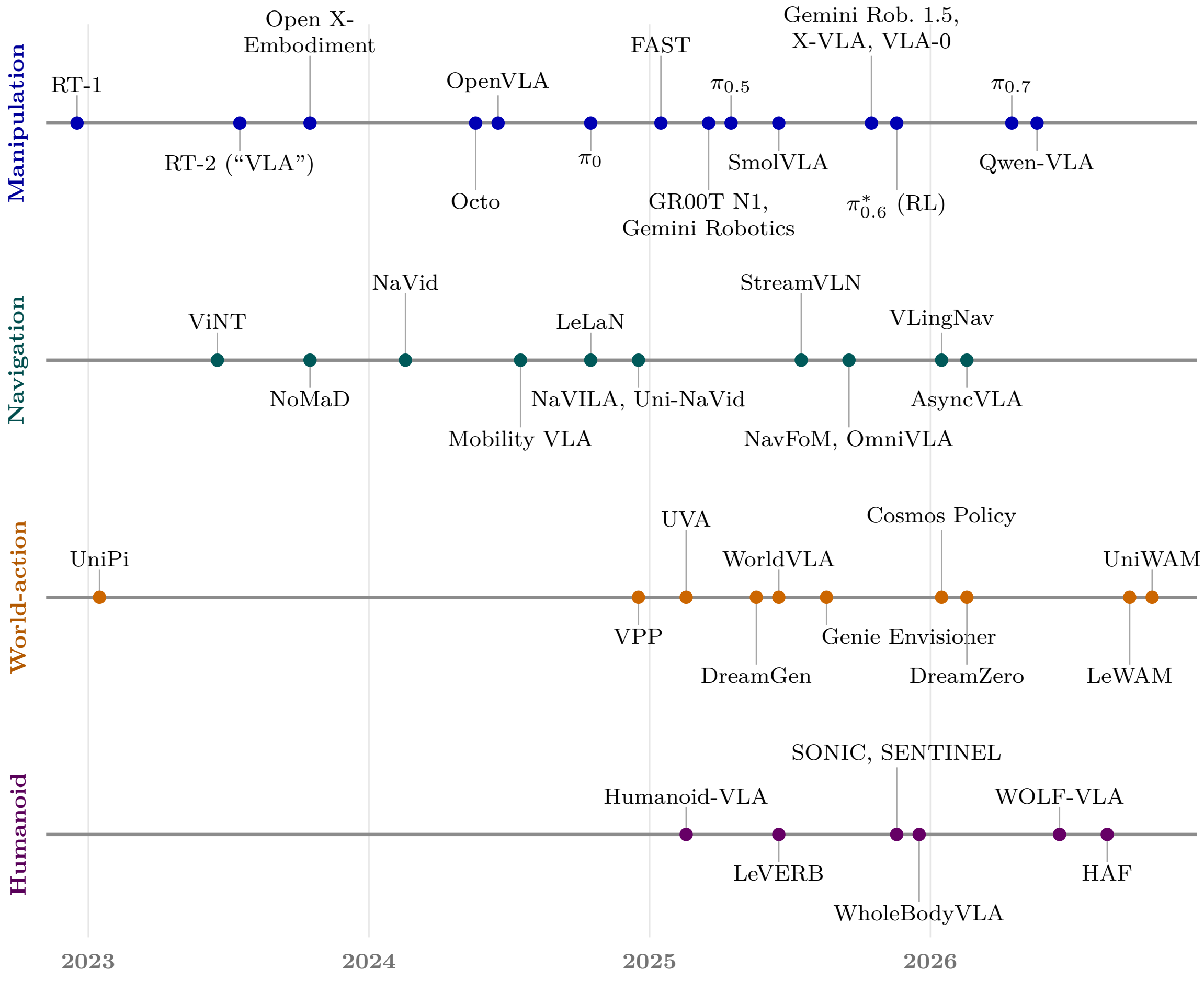
\begin{figure}[t]
\centering
\begin{tikzpicture}[
  x=3.55cm, y=1cm,
  track/.style={line width=1.2pt, draw=black!45},
  dot/.style={circle, inner sep=0pt, minimum size=5pt, draw=white, line width=0.8pt},
  stem/.style={draw=black!35, line width=0.5pt},
  lbl/.style={font=\scriptsize, align=center, inner sep=1pt},
  grid/.style={draw=black!10, line width=0.5pt}]

  \def\xmin{22.85} \def\xmax{26.95}
  % year grid and labels
  \foreach \yr in {23,24,25,26} {
    \draw[grid] (\yr,-10.3) -- (\yr,1.25);
    \node[font=\scriptsize\bfseries, text=black!55] at (\yr,-10.6) {20\yr};
  }

  % \milestone{time}{track y}{level offset}{color}{label}
  \newcommand{\milestone}[5]{
    \draw[stem] (#1,#2) -- (#1,#2+#3);
    \fill[#4] (#1,#2) circle (2.4pt);
    \ifdim #3pt>0pt
      \node[lbl, anchor=south] at (#1,#2+#3) {#5};
    \else
      \node[lbl, anchor=north] at (#1,#2+#3) {#5};
    \fi
  }

  % ---- Manipulation track (y = 0)
  \draw[track] (\xmin,0) -- (\xmax,0);
  \node[rotate=90, font=\scriptsize\bfseries, text=blue!60!black] at (\xmin-0.1,0) {Manipulation};
  \milestone{22.96}{0}{0.35}{blue!70!black}{RT-1}
  \milestone{23.54}{0}{-0.35}{blue!70!black}{RT-2 (``VLA'')}
  \milestone{23.79}{0}{0.85}{blue!70!black}{Open X-\\Embodiment}
  \milestone{24.38}{0}{-0.85}{blue!70!black}{Octo}
  \milestone{24.46}{0}{0.35}{blue!70!black}{OpenVLA}
  \milestone{24.79}{0}{-0.35}{blue!70!black}{$\pi_0$}
  \milestone{25.04}{0}{0.85}{blue!70!black}{FAST}
  \milestone{25.21}{0}{-0.85}{blue!70!black}{GR00T N1,\\Gemini Robotics}
  \milestone{25.29}{0}{0.35}{blue!70!black}{$\pi_{0.5}$}
  \milestone{25.46}{0}{-0.35}{blue!70!black}{SmolVLA}
  \milestone{25.79}{0}{0.85}{blue!70!black}{Gemini Rob.~1.5,\\X-VLA, VLA-0}
  \milestone{25.88}{0}{-0.85}{blue!70!black}{$\pi^*_{0.6}$ (RL)}
  \milestone{26.29}{0}{0.35}{blue!70!black}{$\pi_{0.7}$}
  \milestone{26.38}{0}{-0.35}{blue!70!black}{Qwen-VLA}

  % ---- Navigation track (y = -3)
  \draw[track] (\xmin,-3) -- (\xmax,-3);
  \node[rotate=90, font=\scriptsize\bfseries, text=teal!60!black] at (\xmin-0.1,-3) {Navigation};
  \milestone{23.46}{-3}{0.35}{teal!70!black}{ViNT}
  \milestone{23.79}{-3}{-0.35}{teal!70!black}{NoMaD}
  \milestone{24.13}{-3}{0.85}{teal!70!black}{NaVid}
  \milestone{24.54}{-3}{-0.85}{teal!70!black}{Mobility VLA}
  \milestone{24.79}{-3}{0.35}{teal!70!black}{LeLaN}
  \milestone{24.96}{-3}{-0.35}{teal!70!black}{NaVILA, Uni-NaVid}
  \milestone{25.54}{-3}{0.85}{teal!70!black}{StreamVLN}
  \milestone{25.71}{-3}{-0.85}{teal!70!black}{NavFoM, OmniVLA}
  \milestone{26.04}{-3}{0.35}{teal!70!black}{VLingNav}
  \milestone{26.13}{-3}{-0.35}{teal!70!black}{AsyncVLA}

  % ---- World-action model track (y = -6)
  \draw[track] (\xmin,-6) -- (\xmax,-6);
  \node[rotate=90, font=\scriptsize\bfseries, text=orange!70!black, align=center] at (\xmin-0.1,-6) {World-action};
  \milestone{23.04}{-6}{0.35}{orange!80!black}{UniPi}
  \milestone{24.96}{-6}{-0.35}{orange!80!black}{VPP}
  \milestone{25.13}{-6}{0.85}{orange!80!black}{UVA}
  \milestone{25.38}{-6}{-0.85}{orange!80!black}{DreamGen}
  \milestone{25.46}{-6}{0.35}{orange!80!black}{WorldVLA}
  \draw[stem] (25.63,-6) -- (25.63,-6.35);
  \fill[orange!80!black] (25.63,-6) circle (2.4pt);
  \node[lbl, anchor=north west] at (25.60,-6.35) {Genie Envisioner};
  \milestone{26.04}{-6}{0.85}{orange!80!black}{Cosmos Policy}
  \milestone{26.13}{-6}{-0.85}{orange!80!black}{DreamZero}
  \milestone{26.71}{-6}{-0.85}{orange!80!black}{LeWAM}
  \milestone{26.79}{-6}{0.35}{orange!80!black}{UniWAM}
  % ---- Humanoid whole-body track (y = -9)
  \draw[track] (\xmin,-9) -- (\xmax,-9);
  \node[rotate=90, font=\scriptsize\bfseries, text=violet!70!black] at (\xmin-0.1,-9) {Humanoid};
  \milestone{25.13}{-9}{0.35}{violet!80!black}{Humanoid-VLA}
  \milestone{25.46}{-9}{-0.35}{violet!80!black}{LeVERB}
  \milestone{25.88}{-9}{0.85}{violet!80!black}{SONIC, SENTINEL}
  \milestone{25.96}{-9}{-0.85}{violet!80!black}{WholeBodyVLA}
  \milestone{26.46}{-9}{0.35}{violet!80!black}{WOLF-VLA}
  \milestone{26.63}{-9}{-0.35}{violet!80!black}{HAF}
\end{tikzpicture}
\caption{Milestones in the surveyed literature, by first arXiv release. Manipulation VLAs, navigation VLAs, world-action models and humanoid whole-body control developed largely in parallel. All are discussed in Sections~\ref{sec:manipulation}--\ref{sec:wam}.}
\label{fig:timeline}
\end{figure}

Figure~\ref{fig:timeline} places the main works on a single timeline.

\paragraph{Scope and selection.} We include works that condition a learned policy on vision and language and produce robot actions or action-level intermediates. Candidates were found by starting from widely used VLAs, following their references and the works that cite them, and searching arXiv; this is a structured, not a systematic, review. A work is discussed in the text if it meets at least one of three criteria:
\begin{enumerate}
  \item it introduces an original idea, such as a new action representation, architecture or training signal;
  \item it makes clear progress on a core problem (manipulation, navigation or locomotion), such as a large gain in real-robot success or a new capability;
  \item it gives the community a tool, dataset, benchmark or method that others build on.
\end{enumerate}
Works that mainly add a small performance gain or apply an existing method to a new domain are cited but not discussed. We exclude systems announced without method details; version updates described in release notes are included when the base model's method is published (GR00T~N1.5--N1.7). A small number of non-VLA precursors (RT-1, ACT, Diffusion Policy, PaLM-E, Open X-Embodiment) are included because later VLAs build directly on them. The survey covers 89 works from December 2022 to October 2026: 5 foundations, 38 manipulation VLAs, 12 navigation VLAs, 5 surveys, and 29 works on humanoid control, world-action models and evaluation. Preprints are marked with an asterisk ($^\ast$) in the tables. Training data is taken from each paper's experiment sections; where a quantity is not reported, we say so.

\paragraph{Related surveys.} \citet{surveyvlaembodiedai} gave the first taxonomy of VLAs, with three strands: building blocks, low-level controllers and high-level planners. \citet{surveyvlaconceptsprogress} cover more than 80 models with an emphasis on applications. \citet{surveypurevla} classify over 300 studies into autoregressive, diffusion, reinforcement-based and hybrid paradigms. \citet{reviewvlarealworld} take a full-stack view that includes hardware, data collection and evaluation. \citet{openhelixdualsystem} survey and empirically compare dual-system architectures. More recent surveys take narrower or different cuts. \citet{surveyvlamanipulation} divide VLM-based manipulation VLAs into monolithic models, with single- or dual-system designs, and hierarchical models that plan through interpretable intermediates. \citet{surveyactiontokenization} organise VLAs by the kind of action token they produce, \citet{surveyefficientvla} review efficiency across models, training and data, and \citet{surveyanatomyvla} trace VLAs from their modules through milestones to open challenges. On the world-model side, \citet{surveyworldmodelrobot} review world models for robot learning, \citet{surveywam} survey world-action models, \citet{surveyworldmodelsembodied} grade world models by the capabilities that improve behaviour, \citet{tutorialwam} give a concise tutorial, and \citet{surveyunifiedrobotlearning} connect representation learning, VLAs and world models in one taxonomy. Our survey differs in four ways: it organises the design space along the life cycle of a VLA and places the main models on it, with a practitioner's guide; it treats navigation and humanoid control as lineages alongside manipulation; it compares world-action models with VLAs from the VLA side; and it extracts training data, model size and inference speed from each paper's full text for a quantitative, data-centred comparison.

\paragraph{How to read this survey.} Section~\ref{sec:background} sets out the problem and the precursors of VLAs. Section~\ref{sec:taxonomy} defines the seven life-cycle axes, places the main models on them (Table~\ref{tab:axesgrid}) and collects their deployment figures (Table~\ref{tab:deploy}). Sections~\ref{sec:manipulation}--\ref{sec:humanoid} trace how manipulation, navigation and humanoid VLAs evolved, and Section~\ref{sec:convergence} discusses where these lines meet. Section~\ref{sec:wam} compares VLAs with world-action models, Section~\ref{sec:benchmarks} reviews benchmarks and fair evaluation, Section~\ref{sec:challenges} sets out open challenges and Section~\ref{sec:conclusion} summarises the findings. Appendix~\ref{app:guide} is a practitioner's guide to building a VLA, and Appendix~\ref{app:glossary} explains the terms used. Readers new to VLAs may start with Appendix~\ref{app:glossary}, Figure~\ref{fig:timeline} and Sections~\ref{sec:background}, \ref{sec:manipulation} and~\ref{sec:navigation}; readers building a model with Section~\ref{sec:taxonomy} and Appendix~\ref{app:guide}; and readers looking for research problems with Sections~\ref{sec:benchmarks} and~\ref{sec:challenges}.

\section{Background}
\label{sec:background}

\subsection{Problem formulation}
Following common notation~\citep{memer,pi0}, a language-conditioned robot policy models
\begin{equation}
  \label{eq:policy}
  \pi(A_t \mid o_t), \qquad A_t = [a_t, a_{t+1}, \dots, a_{t+H-1}], \qquad o_t = [I_t, \ell_t, q_t],
\end{equation}
where $I_t$ are images from one or more cameras, $\ell_t$ is the language instruction, $q_t$ is proprioceptive state, and $A_t$ is an \emph{action chunk} of $H$ future actions. Action chunking, introduced with ACT~\citep{actaloha}, reduces compounding error and is now standard in recent manipulation VLAs, although most navigation VLAs still predict one step at a time. A VLA initialises most of $\pi$ from a pretrained VLM. A hierarchical VLA factorises the policy through an intermediate $z_t$:
\begin{equation}
  \pi(A_t \mid o_{0:t}) = \pi_{\text{low}}(A_t \mid I_t, q_t, z_t)\;\pi_{\text{high}}(z_t \mid I_{0:t}, \ell_t),
\end{equation}
where $z_t$ may be a language subtask~\citep{hirobot,memer}, a 2D path~\citep{hamster}, a goal frame~\citep{mobilityvla} or a mid-level motion command~\citep{navila}.

\subsection{Precursors}
\textbf{RT-1}~\citep{rt1} trained a 35M-parameter transformer on about 130k real episodes from 13 robots of a single type. Its ablations showed that data diversity matters more than data volume. \textbf{PaLM-E}~\citep{palme} injected continuous sensor encodings into a 562B-parameter language model as a high-level planner and showed positive transfer from mixing tasks. \textbf{Diffusion Policy}~\citep{diffusionpolicy} cast visuomotor control as conditional denoising and handled multimodal action distributions, with an average improvement of 46.9\% over prior methods. \textbf{ACT}~\citep{actaloha} introduced action chunking on low-cost bimanual hardware and learned fine skills from 50 demonstrations. \textbf{Open X-Embodiment}~\citep{openxembodiment} pooled more than 1M trajectories from 22 embodiments and 21 institutions into a standard format. It became a common pretraining corpus for academic VLAs; most industrial models train mainly on their own data.

\section{Design Axes Along the VLA Life Cycle}
\label{sec:taxonomy}

\begin{figure}[t]
\centering
\def\opt#1#2{$\bullet$\,#1\\[-1pt]\hspace*{0.75em}\parbox[t]{\dimexpr\linewidth-0.75em}{\raggedright\textcolor{black!55}{#2}}\\[2pt]}%
\def\axiscard#1#2#3#4{\begin{minipage}[t][#1][t]{#2}\raggedright\scriptsize
  \textbf{#3}\\[1pt]#4\end{minipage}}%
\resizebox{\linewidth}{!}{\begin{tikzpicture}[
  font=\footnotesize,
  stage/.style={draw=#1!60!black, fill=#1!22, rounded corners=2pt, minimum height=0.7cm,
                font=\bfseries\scriptsize, align=center, inner sep=2pt},
  card/.style={draw=#1!55!black, fill=#1!6, rounded corners=2pt, anchor=north west, inner sep=4pt},
  flow/.style={-{Latex[length=2.2mm]}, very thick, black!60},
  loop/.style={-{Latex[length=2mm]}, thick, black!55, dashed}]

  \node[stage=blue, minimum width=8.08cm, anchor=west] (model) at (0,0) {Model: what it is};
  \node[stage=orange, minimum width=3.78cm, anchor=west] (train) at (8.42,0) {Training: how it learns};
  \node[stage=teal, minimum width=3.78cm, anchor=west] (deploy) at (12.55,0) {Deployment: how it runs};
  \draw[flow] (model.east) -- (train.west);
  \draw[flow] (train.east) -- (deploy.west);

  \node[card=blue] (a) at (0,-0.6) {\axiscard{3.35cm}{3.7cm}{(a) Inputs and memory}{%
    \textit{What does it observe and keep?}\\[3pt]
    \opt{Cameras, language, state}{all VLAs}
    \opt{Extra conditioning}{Hi Robot, $\pi_{0.7}$}
    \opt{Memory of the past}{MemER, StreamVLN}}};
  \node[card=blue] (b) at (4.1,-0.6) {\axiscard{3.35cm}{3.7cm}{(b) Architecture}{%
    \textit{How is the pipeline built?}\\[3pt]
    \opt{Monolithic}{RT-2, OpenVLA, $\pi_0$}
    \opt{Dual system}{GR00T N1, CogACT}
    \opt{Hierarchical}{Hi Robot, NaVILA}}};
  \node[card=blue] (c) at (0,-4.45) {\axiscard{3.35cm}{3.7cm}{(c) Action output}{%
    \textit{How does it produce motor commands?}\\[3pt]
    \opt{Discrete tokens}{RT-2, FAST}
    \opt{Diffusion or flow matching}{$\pi_0$, GR00T N1}
    \opt{Continuous regression}{OpenVLA-OFT}}};
  \node[card=blue] (d) at (4.1,-4.45) {\axiscard{3.35cm}{3.7cm}{(d) Intermediate predictions}{%
    \textit{What else does it predict?}\\[3pt]
    \opt{Reasoning in text}{ECoT, $\pi_{0.5}$}
    \opt{Future observations}{GR-2, VLA-JEPA, DreamZero}}};

  \node[card=orange] (e) at (8.42,-0.6) {\axiscard{3.35cm}{3.5cm}{(e) Pretraining data}{%
    \textit{What does it learn from?}\\[3pt]
    \opt{Robot teleoperation}{$\pi_0$}
    \opt{Cross-embodiment data}{OXE, X-VLA}
    \opt{Web, human video, simulation}{GR00T N1, Qwen-VLA}}};
  \node[card=orange] (f) at (8.42,-4.45) {\axiscard{3.35cm}{3.5cm}{(f) Post-training}{%
    \textit{How is it adapted and improved?}\\[3pt]
    \opt{Fine-tuning recipes}{OpenVLA-OFT}
    \opt{Learning from deployment}{$\pi^*_{0.6}$, VLingNav}}};

  \node[card=teal] (g) at (12.55,-0.6) {\axiscard{7.2cm}{3.5cm}{(g) Real-time execution}{%
    \textit{How does a model running at a few hertz keep pace with the robot?}\\[3pt]
    \opt{Action chunks}{$\pi_0$, ACT}
    \opt{Asynchronous inference}{SmolVLA}
    \opt{Real-time chunking}{RTC}
    \opt{Server--robot split}{AsyncVLA}}};

  \draw[loop] (g.south) -- ++(0,-0.45) -| node[pos=0.25, below, font=\scriptsize\itshape, text=black!65]
    {autonomous experience and corrections feed back into training} (f.south);
\end{tikzpicture}}
\caption{The life-cycle taxonomy of this survey. Seven design axes follow a VLA from model design through training to deployment; each axis poses one question, answered by the main options and example models. Section~\ref{sec:taxonomy} defines the axes and Table~\ref{tab:axesgrid} places 22 models on them.}
\label{fig:lifecycle}
\end{figure}
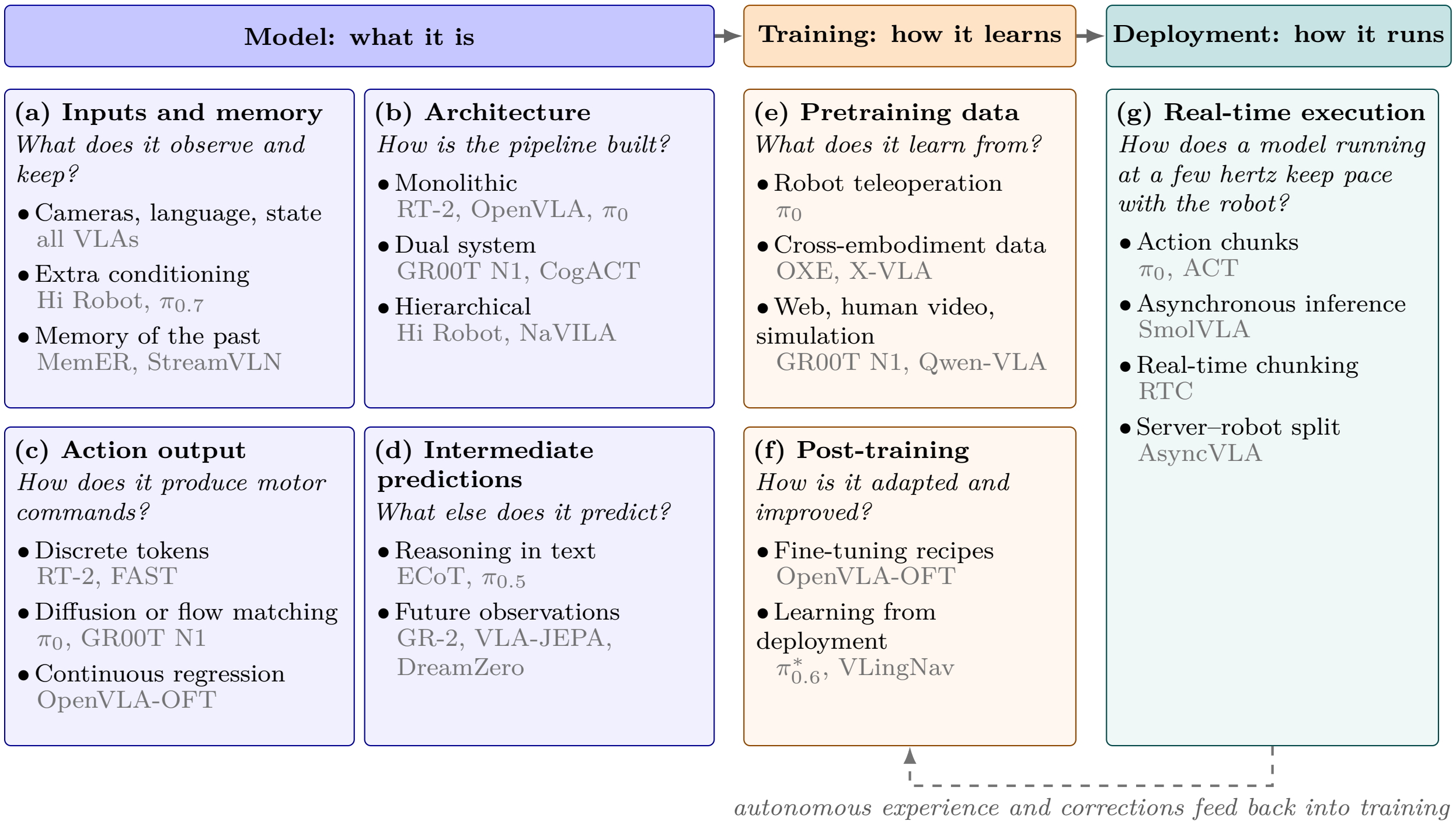

We organise the design space by following a VLA through its life cycle. Four axes describe the model itself: what it observes (Section~\ref{sec:inputs}), how its pipeline is structured (Section~\ref{sec:arch}), how it produces motor commands (Section~\ref{sec:actionout}) and what else it predicts (Section~\ref{sec:intermediate}). Two describe how it learns: the data it is pretrained on (Section~\ref{sec:data}) and how it is adapted afterwards (Section~\ref{sec:posttraining}). The last describes how it runs (Section~\ref{sec:realtime}). Section~\ref{sec:grid} places 22 main models on all seven axes. Figure~\ref{fig:lifecycle} summarises the axes, the question each answers and its main options.

\subsection{Inputs and memory}
\label{sec:inputs}
Following Equation~\eqref{eq:policy}, most VLAs observe one or more camera images, a language instruction and, in manipulation, the robot's proprioceptive state, at the current time step only. Two extensions recur. First, the conditioning has become richer: Hi Robot accepts open-ended prompts and live user corrections~\citep{hirobot}, $\pi_{0.7}$ conditions on episode metadata and subgoal images~\citep{pi07}, and cross-embodiment models add prompts or tokens that identify the robot, dataset or camera view (Section~\ref{sec:data}). Second, some models observe the past.

\textbf{Memory.} Most VLAs are effectively Markovian. Long observation histories are costly to process, and policies trained on them can latch onto cues that do not cause the correct action~\citep{causalconfusion,memer}. MemER~\citep{memer} has a high-level VLM nominate keyframes that are consolidated by 1D clustering into a compact episodic memory, enabling manipulation tasks that need several minutes of recall. StreamVLN~\citep{streamvln} combines a fast sliding-window context with a slow, 3D-aware pruned memory and reuses the transformer's key--value (KV) cache to bound latency. VLingNav maintains a visual-assisted linguistic memory to avoid re-exploring the same places. NaVid~\citep{navid} uses the video history itself as spatio-temporal context.

\subsection{Architecture}
\label{sec:arch}
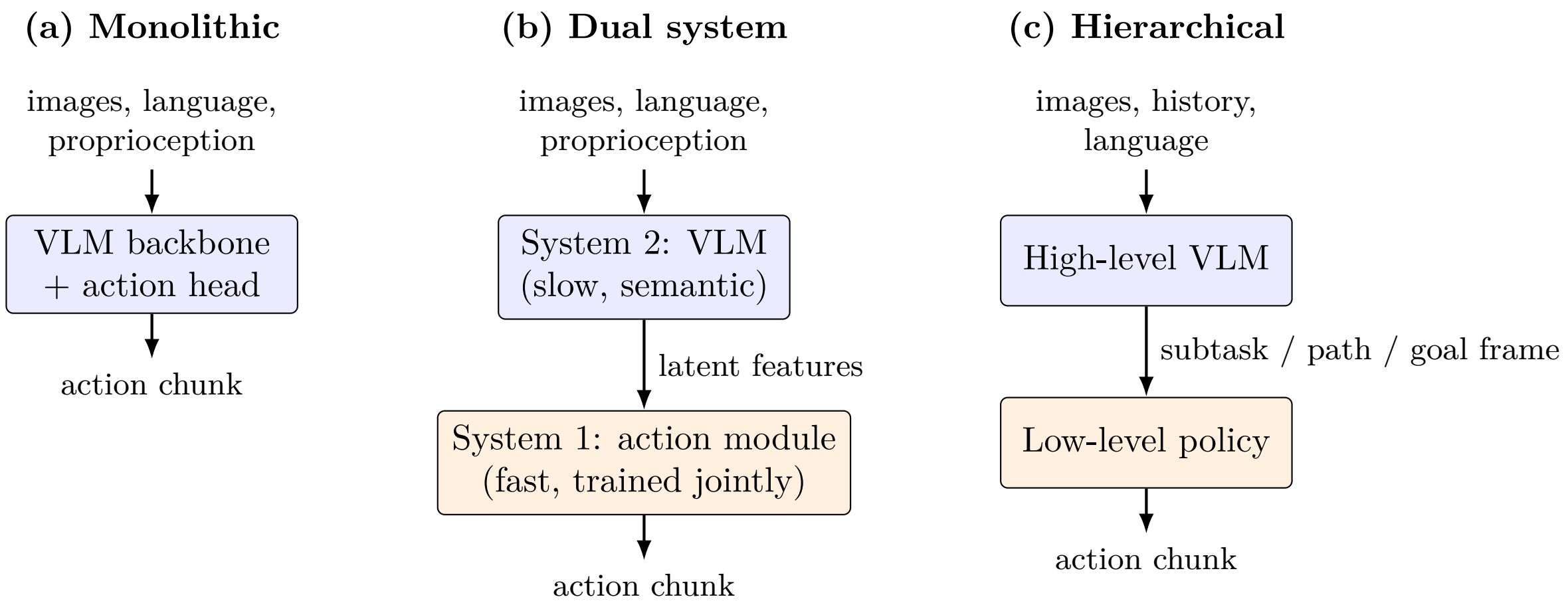
\begin{figure}[t]
\centering
\begin{tikzpicture}[
  font=\small,
  box/.style={draw, rounded corners=2pt, align=center, minimum height=0.9cm, inner sep=4pt},
  vlm/.style={box, fill=blue!8},
  act/.style={box, fill=orange!12},
  io/.style={align=center, font=\footnotesize},
  arr/.style={-{Latex[length=2mm]}, thick}]
  % Monolithic
  \node[io] (i1) at (0,0) {images, language,\\proprioception};
  \node[vlm, below=0.45cm of i1, minimum width=2.9cm] (m1) {VLM backbone\\+ action head};
  \node[io, below=0.45cm of m1] (o1) {action chunk};
  \draw[arr] (i1) -- (m1); \draw[arr] (m1) -- (o1);
  \node[font=\bfseries\small, above=0.15cm of i1] {(a) Monolithic};
  % Dual system
  \node[io] (i2) at (4.9,0) {images, language,\\proprioception};
  \node[vlm, below=0.45cm of i2, minimum width=2.9cm] (s2) {System 2: VLM\\(slow, semantic)};
  \node[act, below=0.9cm of s2, minimum width=2.9cm] (s1) {System 1: action module\\(fast, trained jointly)};
  \node[io, below=0.45cm of s1] (o2) {action chunk};
  \draw[arr] (i2) -- (s2); \draw[arr] (s2) -- node[right, font=\footnotesize]{latent features} (s1); \draw[arr] (s1) -- (o2);
  \node[font=\bfseries\small, above=0.15cm of i2] {(b) Dual system};
  % Hierarchical
  \node[io] (i3) at (9.9,0) {images, history,\\language};
  \node[vlm, below=0.45cm of i3, minimum width=2.9cm] (h3) {High-level VLM};
  \node[act, below=0.9cm of h3, minimum width=2.9cm] (l3) {Low-level policy};
  \node[io, below=0.45cm of l3] (o3) {action chunk};
  \draw[arr] (i3) -- (h3); \draw[arr] (h3) -- node[right, font=\footnotesize]{subtask / path / goal frame} (l3); \draw[arr] (l3) -- (o3);
  \node[font=\bfseries\small, above=0.15cm of i3] {(c) Hierarchical};
\end{tikzpicture}
\caption{Three architectural patterns. (a) One network maps observations to actions (RT-2, OpenVLA, $\pi_0$). (b) A slow VLM reasoner passes latent features to a fast action module trained jointly with it (GR00T~N1, CogACT). (c) The high-level model emits an interpretable intermediate that a separately trained low-level policy executes (Hi Robot, HAMSTER, MemER, Mobility VLA, NaVILA).}
\label{fig:arch}
\end{figure}

Figure~\ref{fig:arch} contrasts three architectural patterns. The split follows \citet{surveyvlamanipulation}, except that we treat dual-system designs as a pattern of their own rather than as a kind of monolithic model.

\textbf{Monolithic} models such as RT-2, OpenVLA and $\pi_0$ map observations directly to actions. We count $\pi_0$ as monolithic even though its action expert has separate weights, because the expert is part of the same transformer and attends to the VLM's tokens at every layer; dual systems instead pass a single latent from one network to another. $\pi_{0.5}$~\citep{pi05} keeps a single network but predicts a text subtask before its actions, an internal hierarchy.

\textbf{Dual-system} models couple a slow VLM (``System~2'') with a fast action module (``System~1'') through latent features and train them jointly: GR00T~N1~\citep{gr00tn1}, CogACT~\citep{cogact}. Gemini Robotics~\citep{geminirobotics} also splits a cloud-hosted VLM backbone from an on-robot action decoder, but does not describe how the two are connected or trained. OpenHelix~\citep{openhelixdualsystem} provides a systematic empirical comparison of these designs.

\textbf{Hierarchical} models make the intermediate explicit and interpretable. Hi Robot~\citep{hirobot} reasons over open-ended prompts and live user corrections to produce language subtasks for a $\pi_0$ controller. HAMSTER~\citep{hamster} predicts a 2D end-effector path that a 3D-aware policy follows; because the path is easy to label, cheap off-domain data becomes usable for training. MolmoAct~\citep{molmoact} produces depth tokens, then an editable trajectory trace, then actions. The hierarchical pattern also lets memory and slow reasoning live outside the fast control loop, as in MemER~\citep{memer}.

\subsection{Action output}
\label{sec:actionout}
How a VLA turns VLM features into continuous robot actions is the most actively explored axis. Table~\ref{tab:actions} summarises the main families.

\begin{table}[t]
\centering\small
\caption{Families of action representation in VLAs.}
\label{tab:actions}
\begin{tabularx}{\linewidth}{L{3.1cm}L{4.2cm}X}
\toprule
Family & Representative works & Trade-off \\
\midrule
Discrete bin tokens & RT-2~\citep{rt2}, OpenVLA~\citep{openvla} & Reuses the language-model head unchanged; per-dimension binning struggles with high-frequency dexterous data and decodes slowly. \\
Compressed tokens & FAST~\citep{fast}, SpatialVLA~\citep{spatialvla} & Compression with the discrete cosine transform (DCT) lets autoregressive VLAs match diffusion VLAs with up to $5\times$ faster training; adaptive grids re-discretise per robot. \\
Plain text numbers & VLA-0~\citep{vla0} & No change to the VLM; reported to outperform several more complex designs on LIBERO~\citep{libero}. \\
Latent actions & LAPA~\citep{lapa}, UniVLA~\citep{univla}, VLA-JEPA~\citep{vlajepa} & Pretrain on action-free human or web video; a small decoder maps latents to real actions. \\
Diffusion head or expert & Octo~\citep{octo}, CogACT~\citep{cogact}, RDT-1B~\citep{rdt1b}, DexVLA~\citep{dexvla}, GR00T~N1~\citep{gr00tn1} & Represents multimodal action distributions; requires iterative sampling. \\
Flow-matching expert & $\pi_0$~\citep{pi0}, $\pi_{0.5}$~\citep{pi05}, SmolVLA~\citep{smolvla}, X-VLA~\citep{xvla} & Smooth high-rate chunks; the current default in frontier systems. \\
Continuous regression & OpenVLA-OFT~\citep{openvlaoft}, VLA-Adapter~\citep{vlaadapter} & Parallel decoding with an L1 loss; fastest inference, well suited to narrow fine-tuning. \\
\bottomrule
\end{tabularx}
\end{table}

Two refinements cut across these families. First, adding an untrained action expert to a pretrained VLM can damage the VLM's semantic knowledge and slow training. \emph{Knowledge insulation}~\citep{pi05knowledgeinsulation} stops gradients flowing from the expert into the backbone while the backbone learns from discrete action tokens. Second, the denoising schedule itself can adapt: AsyncVLA's asynchronous flow matching~\citep{asyncvlaflowmatching}\footnote{Two unrelated 2025--2026 works share the name AsyncVLA. We write AsyncVLA (flow matching) for this manipulation work~\citep{asyncvlaflowmatching} and AsyncVLA (navigation) for the remote-VLA navigation system~\citep{asyncvlanav}.} uses a non-uniform, context-aware schedule and refines low-confidence action tokens before execution.

\subsection{Intermediate predictions}
\label{sec:intermediate}
Besides actions, many VLAs predict outputs that make their decisions explicit or anticipate what will happen next. These take two forms: reasoning in text, and predictions of future observations. In hierarchical systems such outputs are passed from one model to another (Section~\ref{sec:arch}); here we focus on what is predicted.

\textbf{Reasoning before acting.} Embodied chain-of-thought (ECoT)~\citep{ecot} trains a VLA to reason through plans, subtasks, motions, bounding boxes and gripper positions before acting, improving OpenVLA's absolute success rate by 28\% without additional robot data. Gemini Robotics~1.5~\citep{geminirobotics1p5} interleaves natural-language reasoning with actions. In navigation, VLingNav~\citep{vlingnav} uses \emph{adaptive} chain-of-thought, reasoning explicitly only when needed.

\textbf{Predicting the future.} A growing set of VLAs predict future observations as well as actions. 3D-VLA~\citep{threedvla} generates goal images and point clouds with aligned diffusion models. GR-2~\citep{gr2} is pretrained as a video generator before learning actions. VLA-JEPA~\citep{vlajepa}, built on the joint-embedding predictive architecture (JEPA), predicts embeddings of future frames rather than their pixels, so it is not trained to reproduce irrelevant visual detail, and the part of the model that predicts never sees the future frames themselves. $\pi_{0.7}$ uses a lightweight world model to produce subgoal images. These models are a bridge to \emph{world-action models}, which jointly predict future observations and actions; Section~\ref{sec:wam} compares the two families.

\subsection{Pretraining data}
\label{sec:data}
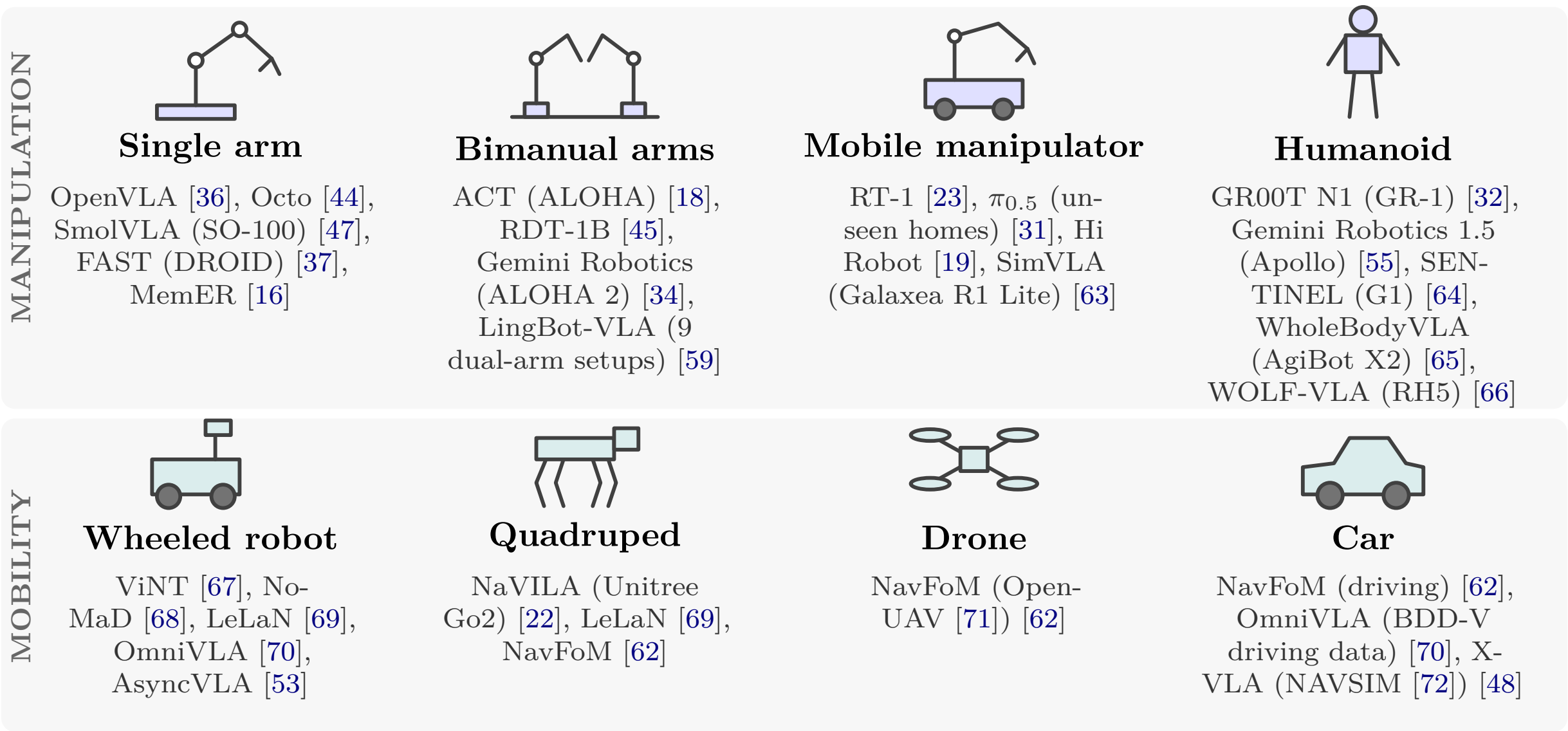
\begin{figure}[t]
\centering
\begin{tikzpicture}[
  font=\small,
  ink/.style={draw=black!75, line width=1.1pt, line cap=round, line join=round},
  body/.style={ink, fill=blue!12},
  nav/.style={ink, fill=teal!14},
  joint/.style={ink, fill=white},
  wheel/.style={ink, fill=black!55},
  label/.style={font=\small\bfseries, align=center},
  examples/.style={font=\scriptsize, align=center, text width=3.3cm, text=black!80, anchor=north},
  band/.style={fill=black!3, rounded corners=4pt}]

  % Row bands
  \fill[band] (-2.0,-0.98) rectangle (14.1,3.15);
  \fill[band] (-2.0,-4.3) rectangle (14.1,-1.08);
  \node[rotate=90, font=\scriptsize\bfseries, text=black!60] at (-1.8,1.3) {MANIPULATION};
  \node[rotate=90, font=\scriptsize\bfseries, text=black!60] at (-1.8,-2.7) {MOBILITY};

  % ---- Single arm
  \begin{scope}[shift={(0,2.0)}]
    \draw[body] (-0.4,0) rectangle (0.4,0.14);
    \draw[ink] (0,0.14) -- (0,0.6) -- (0.45,0.92) -- (0.78,0.62);
    \draw[ink] (0.78,0.62) -- (0.86,0.46); \draw[ink] (0.78,0.62) -- (0.66,0.5);
    \draw[joint] (0,0.6) circle (0.07); \draw[joint] (0.45,0.92) circle (0.07);
  \end{scope}
  \node[label] at (0.15,1.7) {Single arm};
  \node[examples] at (0.15,1.45) {OpenVLA~\citep{openvla}, Octo~\citep{octo}, SmolVLA (SO-100)~\citep{smolvla}, FAST (DROID)~\citep{fast}, MemER~\citep{memer}};

  % ---- Bimanual arms
  \begin{scope}[shift={(4,2.0)}]
    \draw[ink] (-0.75,0.02) -- (0.75,0.02);
    \draw[body] (-0.62,0.02) rectangle (-0.38,0.16);
    \draw[body] (0.38,0.02) rectangle (0.62,0.16);
    \draw[ink] (-0.5,0.16) -- (-0.5,0.62) -- (-0.18,0.86) -- (-0.04,0.6);
    \draw[ink] (0.5,0.16) -- (0.5,0.62) -- (0.18,0.86) -- (0.04,0.6);
    \draw[joint] (-0.5,0.62) circle (0.06); \draw[joint] (0.5,0.62) circle (0.06);
  \end{scope}
  \node[label] at (4,1.7) {Bimanual arms};
  \node[examples] at (4,1.45) {ACT (ALOHA)~\citep{actaloha}, RDT-1B~\citep{rdt1b}, Gemini Robotics (ALOHA~2)~\citep{geminirobotics}, LingBot-VLA (9 dual-arm setups)~\citep{pragmaticvla}};

  % ---- Mobile manipulator
  \begin{scope}[shift={(8,2.0)}]
    \draw[body] (-0.5,0.12) rectangle (0.5,0.4);
    \draw[wheel] (-0.3,0.1) circle (0.1); \draw[wheel] (0.3,0.1) circle (0.1);
    \draw[ink] (-0.2,0.4) -- (-0.2,0.85);
    \draw[ink] (-0.2,0.85) -- (0.25,0.98) -- (0.55,0.72);
    \draw[ink] (0.55,0.72) -- (0.62,0.57); \draw[ink] (0.55,0.72) -- (0.44,0.6);
    \draw[joint] (-0.2,0.85) circle (0.06);
  \end{scope}
  \node[label] at (8,1.7) {Mobile manipulator};
  \node[examples] at (8,1.45) {RT-1~\citep{rt1}, $\pi_{0.5}$ (unseen homes)~\citep{pi05}, Hi Robot~\citep{hirobot}, SimVLA (Galaxea R1 Lite)~\citep{simvla}};

  % ---- Humanoid
  \begin{scope}[shift={(12,2.0)}]
    \draw[body] (0,1.02) circle (0.13);
    \draw[body] (-0.18,0.48) rectangle (0.18,0.86);
    \draw[ink] (-0.18,0.84) -- (-0.36,0.56); \draw[ink] (0.18,0.84) -- (0.36,0.56);
    \draw[ink] (-0.09,0.48) -- (-0.14,0.02); \draw[ink] (0.09,0.48) -- (0.14,0.02);
  \end{scope}
  \node[label] at (12,1.7) {Humanoid};
  \node[examples] at (12,1.45) {GR00T~N1 (GR-1)~\citep{gr00tn1}, Gemini Robotics~1.5 (Apollo)~\citep{geminirobotics1p5}, SENTINEL (G1)~\citep{sentinel}, WholeBodyVLA (AgiBot~X2)~\citep{wholebodyvla}, WOLF-VLA (RH5)~\citep{wolfvla}};

  % ---- Wheeled ground robot
  \begin{scope}[shift={(0,-2.0)}]
    \draw[nav] (-0.45,0.14) rectangle (0.45,0.5);
    \draw[wheel] (-0.28,0.12) circle (0.12); \draw[wheel] (0.28,0.12) circle (0.12);
    \draw[ink] (0.2,0.5) -- (0.2,0.75); \draw[nav] (0.1,0.75) rectangle (0.36,0.9);
  \end{scope}
  \node[label] at (0.15,-2.3) {Wheeled robot};
  \node[examples] at (0.15,-2.55) {ViNT~\citep{vint}, NoMaD~\citep{nomad}, LeLaN~\citep{lelan}, OmniVLA~\citep{omnivlanav}, AsyncVLA~\citep{asyncvlanav}};

  % ---- Quadruped
  \begin{scope}[shift={(4,-2.0)}]
    \draw[nav] (-0.5,0.5) rectangle (0.3,0.72);
    \draw[nav] (0.3,0.6) rectangle (0.55,0.82);
    \draw[ink] (-0.4,0.5) -- (-0.5,0.26) -- (-0.42,0.02);
    \draw[ink] (-0.22,0.5) -- (-0.3,0.26) -- (-0.22,0.02);
    \draw[ink] (0.08,0.5) -- (0.16,0.26) -- (0.08,0.02);
    \draw[ink] (0.24,0.5) -- (0.32,0.26) -- (0.24,0.02);
  \end{scope}
  \node[label] at (4,-2.3) {Quadruped};
  \node[examples] at (4,-2.55) {NaVILA (Unitree Go2)~\citep{navila}, LeLaN~\citep{lelan}, NavFoM~\citep{navfom}};

  % ---- Drone
  \begin{scope}[shift={(8,-2.0)}]
    \draw[ink] (-0.45,0.25) -- (0.45,0.75); \draw[ink] (-0.45,0.75) -- (0.45,0.25);
    \draw[nav] (-0.14,0.38) rectangle (0.14,0.62);
    \foreach \x/\y in {-0.45/0.25, 0.45/0.75, -0.45/0.75, 0.45/0.25}
      \draw[nav] (\x,\y) ellipse (0.2 and 0.06);
  \end{scope}
  \node[label] at (8,-2.3) {Drone};
  \node[examples] at (8,-2.55) {NavFoM (OpenUAV~\citep{openuav})~\citep{navfom}};

  % ---- Car
  \begin{scope}[shift={(12,-2.0)}]
    \draw[nav] (-0.62,0.14) -- (-0.62,0.42) -- (-0.3,0.46) -- (-0.14,0.72) -- (0.26,0.72) -- (0.44,0.46) -- (0.62,0.42) -- (0.62,0.14) -- cycle;
    \draw[wheel] (-0.34,0.13) circle (0.13); \draw[wheel] (0.34,0.13) circle (0.13);
  \end{scope}
  \node[label] at (12,-2.3) {Car};
  \node[examples] at (12,-2.55) {NavFoM (driving)~\citep{navfom}, OmniVLA (BDD-V driving data)~\citep{omnivlanav}, X-VLA (NAVSIM~\citep{navsim})~\citep{xvla}};
\end{tikzpicture}
\caption{Robot bodies in the surveyed work, with models trained or evaluated on each. Several models span more than one body: $\pi_0$ trains on single-arm, dual-arm and mobile manipulators~\citep{pi0}; Gemini Robotics~1.5 transfers skills between ALOHA, a bi-arm Franka and the Apollo humanoid~\citep{geminirobotics1p5}; NavFoM covers quadrupeds, drones, wheeled robots and cars~\citep{navfom}; and Qwen-VLA unifies manipulation and navigation~\citep{qwenvla}.}
\label{fig:embodiments}
\end{figure}

Pretraining data has grown from RT-1's 130k episodes~\citep{rt1}, to OpenVLA's 970k curated Open X-Embodiment episodes~\citep{openvla}, to about 10k hours for $\pi_0$~\citep{pi0}, about 20k hours for LingBot-VLA~\citep{pragmaticvla}, and ``tens of thousands of hours'' for $\pi^*_{0.6}$~\citep{pi06starrecap}. LingBot-VLA reports steady downstream gains as pretraining grows from 3k to 20k hours. Several mechanisms allow heterogeneous action spaces to be trained together:
\begin{itemize}
  \item a physically interpretable unified action space (RDT-1B~\citep{rdt1b});
  \item per-source learnable soft prompts, learned embeddings that tell the model which dataset and robot each example comes from (X-VLA~\citep{xvla});
  \item embodiment-aware text prompts (Qwen-VLA~\citep{qwenvla});
  \item a Motion Transfer mechanism, whose design the paper does not describe (Gemini Robotics 1.5~\citep{geminirobotics1p5});
  \item camera-view and temporal identifier tokens for navigation (NavFoM~\citep{navfom}).
\end{itemize}
Figure~\ref{fig:embodiments} shows the robot bodies these models cover.

\textbf{Co-training} mixes robot actions with other data, most often web visual question answering (VQA) and captioning. RT-2 brought it to VLAs~\citep{rt2}, and PaLM-E, $\pi_{0.5}$, MolmoAct, NaVid and NavFoM follow it~\citep{palme,pi05,molmoact,navid,navfom}. Its value depends on the data: in $\pi_{0.5}$'s ablations, data from other robots and environments was important for performance in unseen homes, whereas web data mainly helped with objects outside the robot data; the web data's effect on overall task success was not statistically significant~\citep{pi05}.

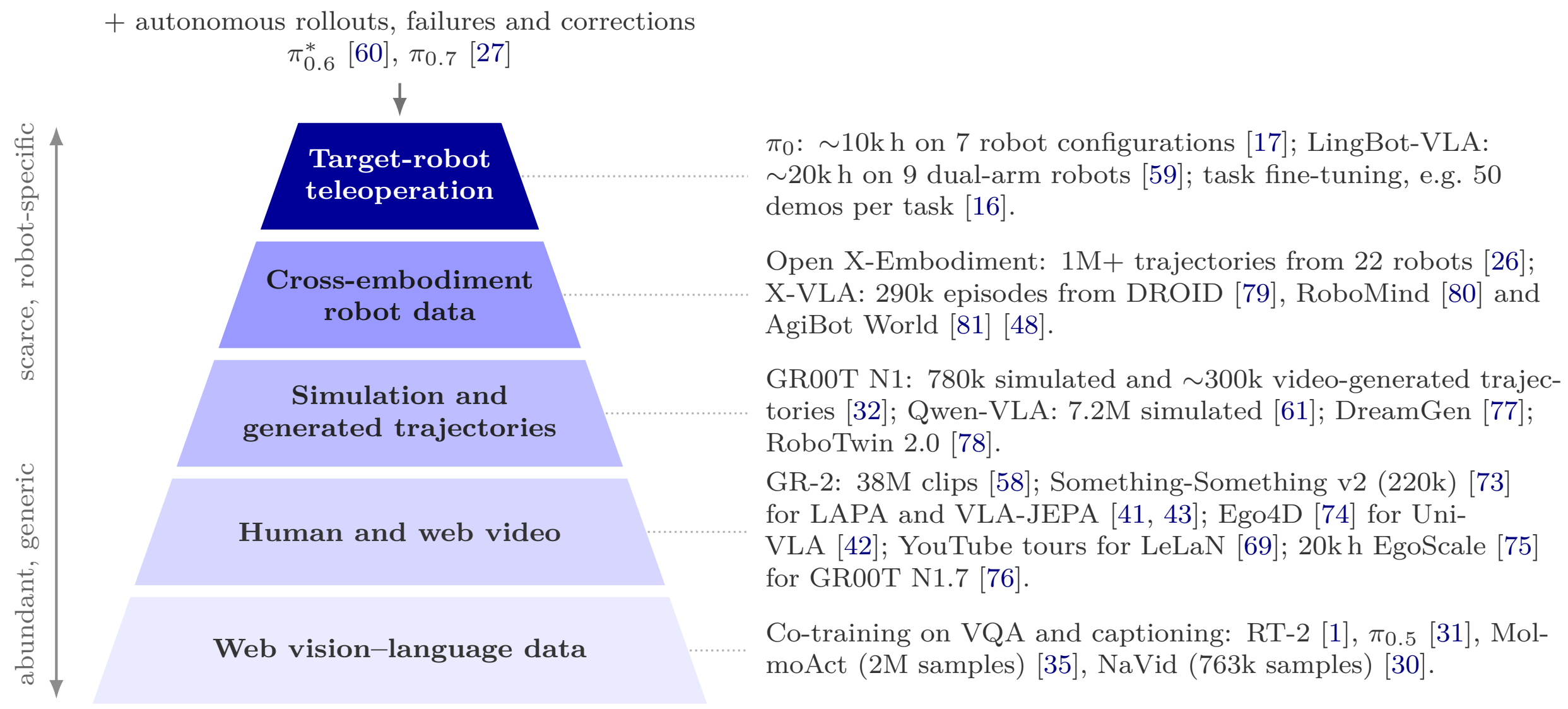
\begin{figure}[t]
\centering
\begin{tikzpicture}[
  y=1.38cm,
  leader/.style={draw=black!30, densely dotted, line width=0.6pt},
  layer/.style={draw=white, line width=1.2pt},
  tiername/.style={font=\scriptsize\bfseries, align=center, text=#1},
  ex/.style={font=\scriptsize, anchor=west, align=left, text width=8.3cm, text=black!80},
  arr/.style={-{Latex[length=2mm]}, line width=0.8pt, draw=black!55}]

  % \tier{y0}{y1}{hw0}{hw1}{fill}
  \newcommand{\tier}[5]{\fill[#5, layer] (3.25-#3,#1) -- (3.25+#3,#1) -- (3.25+#4,#2) -- (3.25-#4,#2) -- cycle;}
  \tier{0}{0.84}{3.25}{2.84}{blue!8}
  \tier{0.9}{1.74}{2.81}{2.40}{blue!16}
  \tier{1.8}{2.64}{2.37}{1.96}{blue!26}
  \tier{2.7}{3.54}{1.93}{1.52}{blue!40}
  \tier{3.6}{4.44}{1.49}{1.08}{blue!60!black}

  \node[tiername=black!80] at (3.25,0.42) {Web vision--language data};
  \node[tiername=black!80] at (3.25,1.32) {Human and web video};
  \node[tiername=black!85] at (3.25,2.22) {Simulation and\\generated trajectories};
  \node[tiername=black!90] at (3.25,3.12) {Cross-embodiment\\robot data};
  \node[tiername=white] at (3.25,4.02) {Target-robot\\teleoperation};

  % leaders and examples
  \foreach \yy/\xx in {0.42/6.29, 1.32/5.85, 2.22/5.41, 3.12/4.97, 4.02/4.53} \draw[leader] (\xx,\yy) -- (6.9,\yy);
  \node[ex] at (6.95,0.42) {Co-training on VQA and captioning: RT-2~\citep{rt2}, $\pi_{0.5}$~\citep{pi05}, MolmoAct (2M samples)~\citep{molmoact}, NaVid (763k samples)~\citep{navid}.};
  \node[ex] at (6.95,1.32) {GR-2: 38M clips~\citep{gr2}; Something-Something~v2 (220k)~\citep{ssv2} for LAPA and VLA-JEPA~\citep{lapa,vlajepa}; Ego4D~\citep{ego4d} for UniVLA~\citep{univla}; YouTube tours for LeLaN~\citep{lelan}; 20k\,h EgoScale~\citep{egoscale} for GR00T~N1.7~\citep{grootn17readme}.};
  \node[ex] at (6.95,2.22) {GR00T~N1: 780k simulated and $\sim$300k video-generated trajectories~\citep{gr00tn1}; Qwen-VLA: 7.2M simulated~\citep{qwenvla}; DreamGen~\citep{dreamgen}; RoboTwin~2.0~\citep{robotwin2}.};
  \node[ex] at (6.95,3.12) {Open X-Embodiment: 1M+ trajectories from 22 robots~\citep{openxembodiment}; X-VLA: 290k episodes from DROID~\citep{droid}, RoboMind~\citep{robomind} and AgiBot World~\citep{agibotworld}~\citep{xvla}.};
  \node[ex] at (6.95,4.02) {$\pi_0$: $\sim$10k\,h on 7 robot configurations~\citep{pi0}; LingBot-VLA: $\sim$20k\,h on 9 dual-arm robots~\citep{pragmaticvla}; task fine-tuning, e.g.\ 50 demos per task~\citep{memer}.};

  % autonomous experience feeding back in
  \node[font=\scriptsize, align=center, text=black!80] (auto) at (3.25,5.05) {+ autonomous rollouts, failures and corrections\\$\pi^*_{0.6}$~\citep{pi06starrecap}, $\pi_{0.7}$~\citep{pi07}};
  \draw[arr] (auto.south) -- (3.25,4.47);

  % side axis
  \draw[{Latex[length=2mm]}-{Latex[length=2mm]}, line width=0.8pt, draw=black!45] (-0.35,0.05) -- (-0.35,4.4);
  \node[rotate=90, font=\scriptsize, text=black!60, anchor=south] at (-0.4,1.0) {abundant, generic};
  \node[rotate=90, font=\scriptsize, text=black!60, anchor=south] at (-0.4,3.45) {scarce, robot-specific};
\end{tikzpicture}
\caption{Sources of VLA training data, from abundant but generic (bottom) to scarce but specific to the target robot (top), with figures reported in the surveyed papers. Autonomous experience is a growing source that feeds back into the top layer. The layering follows the spirit of the data pyramid used to train GR00T~N1~\citep{gr00tn1}.}
\label{fig:datasources}
\end{figure}

\textbf{Beyond teleoperation.} Training data increasingly comes from other sources (Figure~\ref{fig:datasources}):
\begin{itemize}
  \item human and web video: 38M clips for GR-2~\citep{gr2}, Something-Something~v2~\citep{ssv2} for LAPA and VLA-JEPA~\citep{lapa,vlajepa}, Ego4D~\citep{ego4d} for UniVLA~\citep{univla}, and YouTube tours for LeLaN~\citep{lelan};
  \item simulation: 780k trajectories for GR00T~N1~\citep{gr00tn1} and 7.2M for Qwen-VLA~\citep{qwenvla};
  \item video-generated ``neural trajectories'' (GR00T~N1);
  \item failed and autonomous episodes, made usable by telling the policy how good each episode or action was, through metadata or an advantage estimate~\citep{pi06starrecap,pi07}.
\end{itemize}

\subsection{Post-training}
\label{sec:posttraining}
Supervised fine-tuning on the target robot is the default. OpenVLA-OFT~\citep{openvlaoft} distils a recipe of parallel decoding, action chunking, continuous actions and an L1 loss. It raises OpenVLA's LIBERO average (Section~\ref{sec:benchmarks}) from 76.5\% to 97.1\% and its throughput by $26\times$. Reinforcement learning from deployment has recently become practical at scale. RECAP~\citep{pi06starrecap} trains a value function and a policy conditioned on the advantage, that is, on how much better than expected each action turned out, using demonstrations, autonomous rollouts and expert interventions together. It was applied to laundry folding, espresso making and box assembly; on the hardest of these tasks it more than doubles throughput and roughly halves the failure rate. VLingNav adds an online, expert-guided RL stage to navigation. $\pi_{0.7}$~\citep{pi07} instead conditions on episode metadata and subgoal images, so lower-quality data improves the model rather than degrading it.

\subsection{Real-time execution}
\label{sec:realtime}
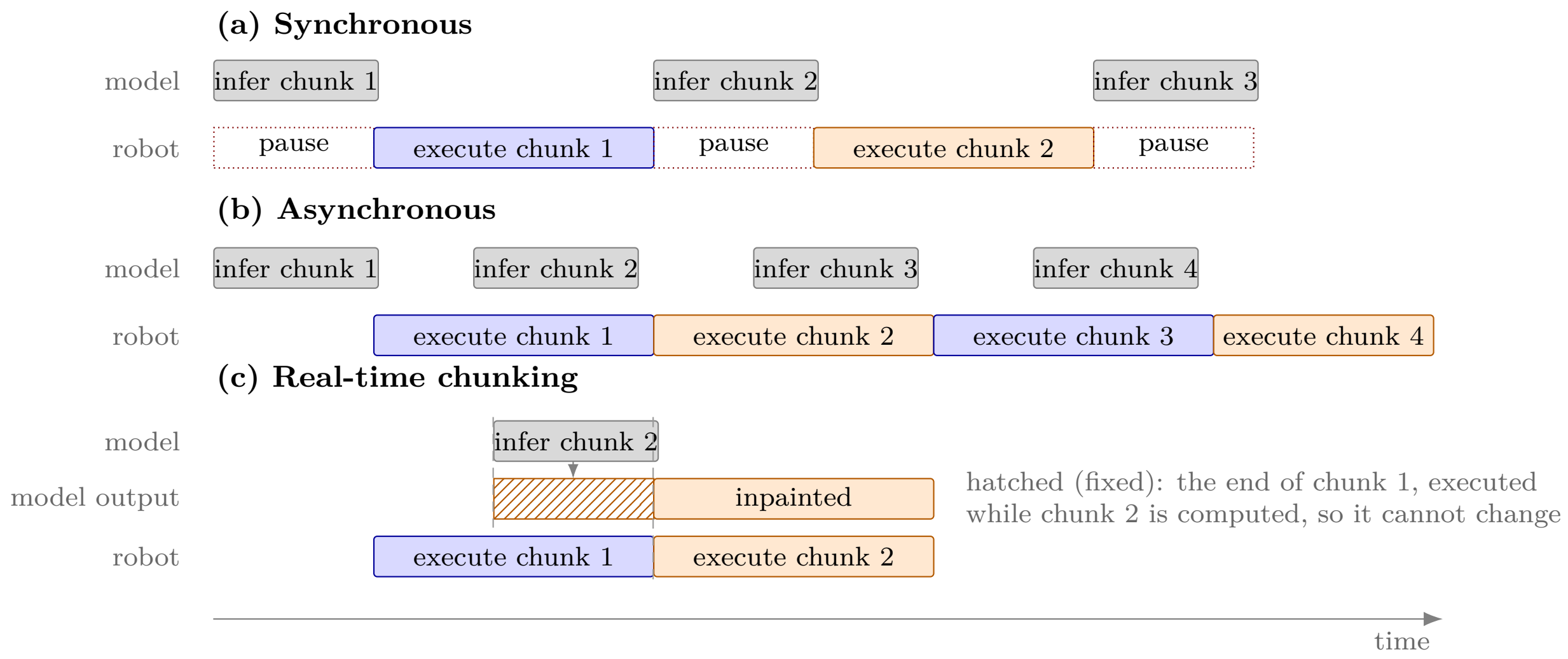
\begin{figure}[t]
\centering
\begin{tikzpicture}[x=1.04cm,
  font=\footnotesize,
  infer/.style={draw=black!50, fill=black!15, rounded corners=1pt, minimum height=0.42cm, inner sep=0pt},
  exA/.style={draw=blue!60!black, fill=blue!15, rounded corners=1pt, minimum height=0.42cm, inner sep=0pt},
  exB/.style={draw=orange!70!black, fill=orange!18, rounded corners=1pt, minimum height=0.42cm, inner sep=0pt},
  frozen/.style={draw=orange!70!black, pattern=north east lines, pattern color=orange!70!black, minimum height=0.42cm, inner sep=0pt},
  idle/.style={draw=red!50!black, densely dotted, minimum height=0.42cm, inner sep=0pt},
  rowlbl/.style={anchor=south west, font=\footnotesize\bfseries},
  tag/.style={font=\scriptsize, text=black!60}]

  % \bar{x0}{x1}{y}{style}{text}
  \newcommand{\seg}[5]{\node[#4, minimum width=(#2-#1)*1.04cm, anchor=west] at (#1,#3) {\scriptsize #5};}

  % axis
  \draw[-{Latex[length=2mm]}, black!50] (0,-5.25) -- (12.3,-5.25) node[below left, tag] {time};

  % (a) synchronous
  \node[rowlbl] at (-0.1,0.62) {(a) Synchronous};
  \node[tag, anchor=east] at (-0.2,-0.35) {robot};
  \node[tag, anchor=east] at (-0.2,0.35) {model};
  \seg{0}{1.6}{0.35}{infer}{infer chunk 1}
  \seg{0}{1.6}{-0.35}{idle}{pause}
  \seg{1.6}{4.4}{-0.35}{exA}{execute chunk 1}
  \seg{4.4}{6.0}{0.35}{infer}{infer chunk 2}
  \seg{4.4}{6.0}{-0.35}{idle}{pause}
  \seg{6.0}{8.8}{-0.35}{exB}{execute chunk 2}
  \seg{8.8}{10.4}{0.35}{infer}{infer chunk 3}
  \seg{8.8}{10.4}{-0.35}{idle}{pause}

  % (b) asynchronous
  \node[rowlbl] at (-0.1,-1.3) {(b) Asynchronous};
  \node[tag, anchor=east] at (-0.2,-2.3) {robot};
  \node[tag, anchor=east] at (-0.2,-1.6) {model};
  \seg{0}{1.6}{-1.6}{infer}{infer chunk 1}
  \seg{2.6}{4.2}{-1.6}{infer}{infer chunk 2}
  \seg{5.4}{7.0}{-1.6}{infer}{infer chunk 3}
  \seg{8.2}{9.8}{-1.6}{infer}{infer chunk 4}
  \seg{1.6}{4.4}{-2.3}{exA}{execute chunk 1}
  \seg{4.4}{7.2}{-2.3}{exB}{execute chunk 2}
  \seg{7.2}{10.0}{-2.3}{exA}{execute chunk 3}
  \seg{10.0}{12.2}{-2.3}{exB}{execute chunk 4}

  % (c) real-time chunking
  \node[rowlbl] at (-0.1,-3.05) {(c) Real-time chunking};
  \node[tag, anchor=east] at (-0.2,-3.4) {model};
  \node[tag, anchor=east] at (-0.2,-4.0) {model output};
  \node[tag, anchor=east] at (-0.2,-4.6) {robot};
  \seg{2.8}{4.4}{-3.4}{infer}{infer chunk 2}
  \seg{2.8}{4.4}{-4.0}{frozen}{}
  \seg{4.4}{7.2}{-4.0}{exB}{inpainted}
  \draw[-{Latex[length=1.5mm]}, black!50] (3.6,-3.61) -- (3.6,-3.79);
  \seg{1.6}{4.4}{-4.6}{exA}{execute chunk 1}
  \seg{4.4}{7.2}{-4.6}{exB}{execute chunk 2}
  \draw[densely dashed, black!40] (2.8,-3.15) -- (2.8,-4.3);
  \draw[densely dashed, black!40] (4.4,-3.15) -- (4.4,-4.85);
  \node[tag, anchor=west, align=left] at (7.4,-4.0) {hatched (fixed): the end of chunk 1, executed\\while chunk 2 is computed, so it cannot change};
\end{tikzpicture}
\caption{Executing action chunks. (a) A synchronous policy stops the robot while the model computes the next chunk, so the robot pauses at every chunk boundary. (b) Asynchronous execution computes the next chunk while the current one runs, as in SmolVLA's inference stack~\citep{smolvla} and the remote-VLA and edge-adapter split of AsyncVLA (navigation)~\citep{asyncvlanav}; but if the new chunk plans a different motion, the robot jumps when it switches, causing jerky movement. (c) Real-time chunking~\citep{realtimechunking} prevents this. The robot carries on executing chunk~1 while the model computes chunk~2; those actions cannot be changed, so the model keeps them fixed (hatched) and predicts the rest of chunk~2 (inpainted), guided by the remainder of chunk~1, so that it continues smoothly.}
\label{fig:execution}
\end{figure}

Large VLAs run at a few hertz, which leads to pauses or jerky transitions at chunk boundaries (Figure~\ref{fig:execution}). SmolVLA~\citep{smolvla} decouples action prediction from execution with an asynchronous inference stack. Real-Time Chunking~\citep{realtimechunking} overlaps the two: while one chunk executes, it plans the next, keeping fixed the actions that will run before the new chunk is ready and filling in the rest so the two chunks join smoothly. It runs at inference time on existing diffusion and flow-matching policies, without retraining. AsyncVLA (navigation)~\citep{asyncvlanav} runs a large VLA on a remote workstation and a lightweight edge adapter on the robot. It tolerates communication delays of up to 6~seconds; among pedestrians it reaches the goal without collision in 85\% of trials, which its authors report as 40\% above the strongest baseline. Hierarchical systems often run the two levels at different rates: MemER runs its high-level policy at about 1~Hz and its low-level policy at about 2~Hz~\citep{memer}.

\textbf{Where VLAs run.} Table~\ref{tab:deploy} collects the model size, inference speed and hardware that the main works report. Three patterns stand out. Every model that names its hardware is run on a single desktop or data-centre GPU (RTX~4090 or 5090, A6000, A100, L40, H100) or on cloud TPUs; embedded robot hardware appears only for the edge part of a split system, such as AsyncVLA's adapter on a Jetson Orin. Navigation VLAs built on 7--8B video VLMs are usually served from a remote RTX~4090 or A100 over a network. And speed is reported in incompatible forms (throughput, latency per chunk, frames per second, effective control rate) on different GPUs, so the figures in the table cannot be ranked directly; VLA-Perf~\citep{vlaperf} addresses this with an analytical latency model (Section~\ref{sec:benchmarks}).

\begin{table}[!htb]
\centering\footnotesize
\setlength{\tabcolsep}{4pt}
\caption{Model size, reported inference speed and hardware for the main VLAs, as stated in each paper. Speeds are in the paper's own units and are not directly comparable. n.r.: not reported.}
\label{tab:deploy}
\begin{tabularx}{\linewidth}{L{2.3cm}L{1.3cm}XL{3.6cm}}
\toprule
Model & Params & Reported inference speed & Hardware \\
\midrule
RT-2~\citep{rt2} & 55B (5B) & 1--3\,Hz (5B: $\sim$5\,Hz) & Multi-TPU cloud service, queried over the network \\
OpenVLA~\citep{openvla} & 7B & $\sim$6\,Hz; 15\,GB in bf16 & One RTX~4090 \\
TinyVLA~\citep{tinyvla} & 1B & 14\,ms per action prediction (OpenVLA-7B: 292\,ms) & One A6000 \\
$\pi_0$~\citep{pi0} & 3.3B & 73\,ms per chunk on board, 86\,ms off board; control up to 50\,Hz & One RTX~4090; off-board over Wi-Fi for the mobile robot \\
OpenVLA-OFT~\citep{openvlaoft} & 7B & 71.4\,Hz throughput, 0.11\,s latency (LIBERO); real bimanual control at 25\,Hz & One A100 \\
GR00T~N1\npr~\citep{gr00tn1} & 2.2B & 63.9\,ms per 16-action chunk; VLM at 10\,Hz, actions at 120\,Hz & One L40 \\
Gemini Robotics\npr~\citep{geminirobotics} & n.r. & $\sim$250\,ms end to end; 50\,Hz effective control & Cloud backbone ($<$160\,ms) plus on-robot decoder \\
SmolVLA\npr~\citep{smolvla} & 0.45B & n.r.; asynchronous inference stack & Designed for consumer GPUs and CPUs \\
VLA-Adapter~\citep{vlaadapter} & 0.5B & 219.2\,Hz throughput, 0.04\,s latency (LIBERO) & H100 \\
VLA-0\npr~\citep{vla0} & 3B & 4\,Hz & Desktop with one RTX~5090 \\
\midrule
NaVid~\citep{navid} & 7B LLM & n.r.; authors note long latency & Remote A100 server, over the internet \\
NaVILA~\citep{navila} & 8B & $\sim$1 frame per second & One RTX~4090 \\
StreamVLN~\citep{streamvln} & n.r. & n.r.; KV-cache reuse for low latency & Remote RTX~4090 workstation \\
NavFoM~\citep{navfom} & n.r. & n.r.; token budget keeps speed stable & Remote RTX~4090 server \\
OmniVLA~\citep{omnivlanav} & 7B & 8-action chunks at 3\,Hz & Local PC with one RTX~4090 \\
VLingNav~\citep{vlingnav} & 7B base & $<$300\,ms per step over 500 frames, plus $\sim$100\,ms network & Remote RTX~4090 server \\
AsyncVLA (navi\-gation)\npr~\citep{asyncvlanav} & n.r. & 3\,Hz control; tolerates 6\,s delay & VLA on a remote RTX~4090; edge adapter on a Jetson Orin \\
\bottomrule
\end{tabularx}
\end{table}

\subsection{Placing models on the axes}
\label{sec:grid}
Table~\ref{tab:axesgrid} places 22 main models on the seven axes, in life-cycle order. A dash means the work makes no distinctive choice on that axis, not that the axis is absent. Read by column, the grid shows where effort has gone: action output and pretraining data differ for every model, nine models extend their inputs and ten predict something besides actions, but only seven make a distinctive post-training choice and only two, VLA-JEPA and $\pi_{0.7}$, predict future observations. Navigation models stand out on inputs: five of the seven add history or extra views, because navigation depends on where the robot has been. Read by row, the grid shows design choices accumulating: $\pi_{0.7}$ makes a distinctive choice on all seven axes, while RT-2 and OpenVLA do so on four.

\begin{sidewaystable}
\centering\scriptsize
\setlength{\tabcolsep}{3pt}
\renewcommand{\arraystretch}{0.95}
\caption{Main VLAs placed on the seven design axes of Section~\ref{sec:taxonomy}, grouped by life-cycle stage. --: no distinctive choice on this axis. FM: flow matching; FT: fine-tuning; OXE: Open X-Embodiment. $^\ast$Preprint.}
\label{tab:axesgrid}
\begin{tabular}{L{2.0cm}L{2.4cm}L{2.0cm}L{2.2cm}L{2.4cm}L{3.3cm}L{2.3cm}L{2.4cm}}
\toprule
 & \multicolumn{4}{c}{Model} & \multicolumn{2}{c}{Training} & Deployment \\
\cmidrule(lr){2-5}\cmidrule(lr){6-7}\cmidrule(lr){8-8}
Model & Inputs and memory & Architecture & Action output & Intermediate predictions & Pretraining data & Post-training & Real-time execution \\
\midrule
\multicolumn{8}{l}{\textit{Manipulation}} \\
RT-2 & -- & Monolithic & Discrete bin tokens & -- & One robot type; web VQA co-training & -- & Cloud TPUs, 1--3\,Hz \\
OpenVLA & -- & Monolithic & Discrete bin tokens & -- & 970k OXE episodes & LoRA FT, quantisation & -- \\
ECoT & -- & Monolithic & Discrete bin tokens & Chain-of-thought: plans, subtasks, boxes & BridgeData V2 + synthetic reasoning & -- & -- \\
$\pi_0$ & -- & Monolithic & FM expert & -- & $\sim$10k\,h, 7 robot configs + OXE & Task FT & Chunks up to 50\,Hz \\
OpenVLA-OFT & -- & Monolithic & Continuous regression & -- & FT only & Parallel decoding, chunking, L1 loss & Parallel decoding \\
GR00T N1\npr & -- & Dual system & FM diffusion transformer & -- & Real humanoid, human video, sim, generated trajectories & -- & VLM 10\,Hz, actions 120\,Hz \\
Hi Robot & Open-ended prompts, live corrections & Hierarchical & FM ($\pi_0$ low level) & Language subtasks & Teleop + synthetic prompts & -- & -- \\
$\pi_{0.5}$ & -- & Monolithic & FM expert & Text subtask & Co-training: homes, cross-embodiment, web & -- & -- \\
SmolVLA\npr & -- & Monolithic & FM expert & -- & 22.9k community episodes & -- & Asynchronous inference \\
MemER & Keyframe episodic memory & Hierarchical & FM ($\pi_{0.5}$ low level) & Language subtasks & 50 demos per task & -- & High level $\sim$1\,Hz, low $\sim$2\,Hz \\
X-VLA & -- & Monolithic & FM transformer & -- & 290k episodes; per-source soft prompts & -- & -- \\
$\pi^*_{0.6}$ & -- & Monolithic & FM expert + FAST tokens & Text subtask & Demos, autonomous rollouts, corrections & RECAP offline RL & -- \\
VLA-JEPA & -- & Monolithic & Latent actions & Future latent embeddings & DROID + human video & -- & -- \\
$\pi_{0.7}$\npr & Episode metadata, subgoal images & VLA with high-level policy and world model & FM expert & Subgoal images from a world model & Demos incl.\ failures, human video & Metadata conditioning & Acts while next subgoal is generated \\
Qwen-VLA\npr & Embodiment-aware prompts & Monolithic & FM diffusion transformer & -- & Real, human, 7.2M sim, VLN & -- & -- \\
\midrule
\multicolumn{8}{l}{\textit{Navigation}} \\
NaVid & Monocular video history & Monolithic & Next-step actions as text & -- & 510k VLN-CE + 763k web & -- & Remote A100 \\
Mobility VLA & Long-context tour video & Hierarchical & Topological-graph controller & Goal frame & No policy training & -- & -- \\
NaVILA & -- & Hierarchical & RL locomotion policy & Mid-level language commands & R2R-CE, RxR-CE, touring video, VQA & -- & $\sim$1 frame per second \\
StreamVLN & Slow-fast memory & Monolithic & Next-step actions & -- & MP3D, ScaleVLN, video QA & DAgger & KV-cache reuse \\
NavFoM & Camera-view and time tokens & Monolithic & Waypoint trajectory & -- & 8.02M samples, 4 body types & -- & Token budget \\
VLingNav & Linguistic memory & Monolithic & Continuous regression & Adaptive chain-of-thought & 4.5M samples & Online RL & $<$300\,ms per step \\
AsyncVLA (navigation)\npr & -- & Remote VLA + on-robot adapter & VLA tokens refined by edge adapter & -- & GNM, SACSoN, LeLaN & -- & Tolerates 6\,s delay \\
\bottomrule
\end{tabular}
\end{sidewaystable}

\section{Manipulation VLAs}
\label{sec:manipulation}

\begin{figure}[t]
\centering
\includegraphics[width=\linewidth]{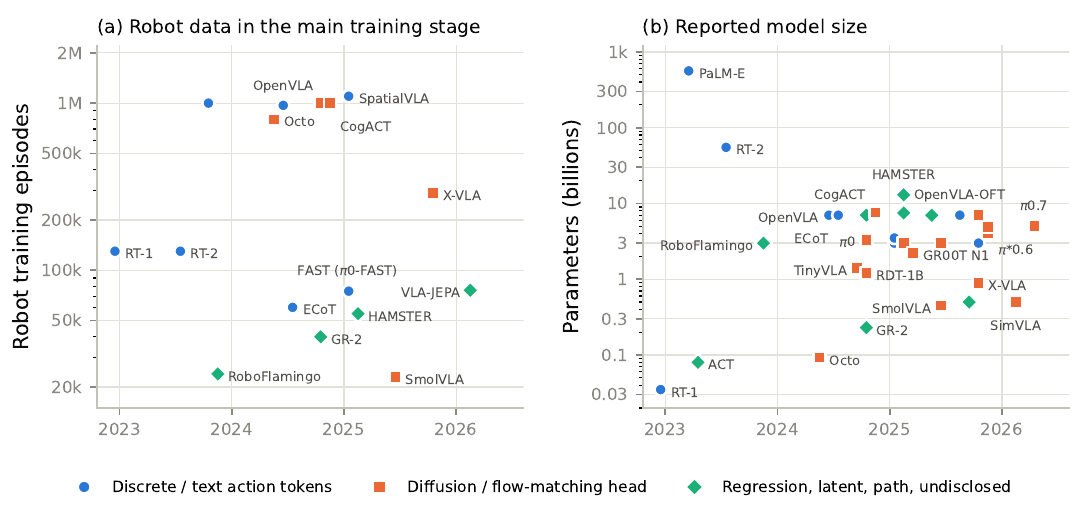}
\caption{Manipulation VLAs over time. (a) Robot episodes in each model's main training stage, for the 16 works that report a single episode count; Open X-Embodiment (unlabelled, late 2023) is the dataset itself. Models reporting only hours, such as $\pi_0$ ($\sim$10k\,h) and LingBot-VLA ($\sim$20k\,h), cannot be placed on this axis; models with no robot pretraining (Diffusion Policy, ACT, TinyVLA, VLA-Adapter, VLA-0) are omitted, as is SimVLA, whose benchmark models use no robot pretraining but whose real-robot model trains on about 500 hours. (b) Reported parameter count for the 32 works that state one; for hierarchical systems the value is the high-level model. Labels are omitted where they would overlap; every point appears in Table~\ref{tab:manip}.}
\label{fig:scaling}
\end{figure}

Table~\ref{tab:manip} lists the 43 manipulation works with their training data and key ideas. Figure~\ref{fig:scaling} plots the subset that reports comparable quantities. We read the history in four phases.

\textbf{2023: emergence.} RT-2~\citep{rt2} established the recipe of actions as tokens inside a large VLM, co-fine-tuned with web data. RoboFlamingo~\citep{roboflamingo} showed that an open VLM with a light policy head could compete on CALVIN~\citep{calvin}.

\textbf{2024: open generalists.} Octo~\citep{octo} and OpenVLA~\citep{openvla} released open models pretrained on 800k and 970k Open X-Embodiment episodes; OpenVLA outperformed the 55B RT-2-X with $7\times$ fewer parameters. ECoT, TinyVLA, RDT-1B, LAPA and CogACT explored reasoning, compactness, bimanual diffusion, latent actions and specialised action modules. $\pi_0$~\citep{pi0} combined a VLM based on PaliGemma~\citep{paligemma} with a flow-matching action expert and about 10k hours of dexterous multi-robot data.

\textbf{2025: efficiency, hierarchy and generalisation.} FAST made autoregressive VLAs competitive. OpenVLA-OFT made fine-tuning efficient. GR00T~N1 released open weights for a dual-system humanoid model, and Gemini Robotics brought Gemini~\citep{gemini} to robot control. $\pi_{0.5}$~\citep{pi05} demonstrated end-to-end long-horizon cleaning in homes never seen during training. At the efficient end, SmolVLA, VLA-Adapter and VLA-0 showed strong results at sub-billion scale or without robot pretraining. MemER added minutes-long memory, and $\pi^*_{0.6}$ brought RL from deployment.

\textbf{2026: steerability and unification.} LingBot-VLA scaled open real-robot data to about 20k hours. VLA-JEPA brought latent world-model pretraining, and SimVLA~\citep{simvla} showed that a minimal 0.5B design with a standardised training recipe matches far larger models, arguing that many reported gains are hard to separate from recipe differences. $\pi_{0.7}$~\citep{pi07} made a generalist steerable through rich context and reported zero-shot cross-embodiment laundry folding. Qwen-VLA~\citep{qwenvla} trained one model for manipulation, navigation and trajectory prediction.

\textbf{Open model lines released without papers.} Some of the most widely used open VLAs now evolve through code releases rather than papers. NVIDIA's GR00T line is the clearest case: only N1 has a paper~\citep{gr00tn1}. N1.5, N1.6 and N1.7 are documented in the release notes of NVIDIA's open-source repository~\citep{grootn15readme,grootn16readme,grootn17readme} (Table~\ref{tab:groot}). Across the four releases the line moves from a frozen VLM to partial fine-tuning of a reasoning-oriented VLM, from absolute to relative actions, and from robot-only data towards large-scale human video. Because these releases have no papers, their data and evaluations cannot be checked to the standard of the works in Table~\ref{tab:manip}.

\begin{table}[!htb]
\centering\small
\caption{The GR00T N1 model line. N1 is described in its paper; later versions in NVIDIA's repository release notes. All four are non-peer-reviewed.}
\label{tab:groot}
\begin{tabularx}{\linewidth}{L{1.25cm}L{3.0cm}X}
\toprule
Version & Vision-language backbone & Main changes \\
\midrule
N1~\citep{gr00tn1} & Eagle-2 VLM~\citep{eagle2} & Dual system: VLM plus a flow-matching diffusion-transformer (DiT) action head, 2.2B parameters; trained on real humanoid data, human video, simulation and generated ``neural trajectories''. \\
N1.5~\citep{grootn15readme} & Eagle~2.5~\citep{eagle25}, frozen in pretraining and fine-tuning & Simplified multilayer-perceptron (MLP) adapter; FLARE future-latent alignment objective alongside flow matching, for learning from egocentric human video; DreamGen trajectories; language following on Fourier GR-1 humanoid tasks reported at 93.3\% versus 46.6\% for N1; new heads for single-arm end-effector control and AgiBot Genie-1. \\
N1.6~\citep{grootn16readme} & Internal Cosmos-Reason-2B variant~\citep{cosmosreason1}\textsuperscript{a}; native aspect ratio & DiT doubled to 32 layers; adapter removed and the top 4 VLM layers unfrozen; state-relative action chunks; several thousand added hours of teleoperation (YAM bimanual arms, AgiBot Genie-1, simulated Galaxea R1 Pro, Unitree G1 loco-manipulation), for 10k+ hours of robot data in total. \\
N1.7~\citep{grootn17readme} & Cosmos-Reason2-2B (Qwen3-VL architecture~\citep{qwen3vl}) & 3B parameters; relative end-effector action space shared by human and robot data; 20k hours of EgoScale~\citep{egoscale} human video in pretraining; DiT reduced to 16 layers; state/action dimensions 29$\rightarrow$132 and action horizon 16$\rightarrow$40; dataset-mixture weighting; ONNX/TensorRT export; Apache-2.0 licence. Reported as comparable to N1.6 with better generalisation and language following. \\
\bottomrule
\end{tabularx}

\smallskip
\raggedright\footnotesize\textsuperscript{a}The N1.6 release notes describe an internal Cosmos-Reason-2B variant, while the N1.7 changelog names the N1.6 backbone \texttt{nvidia/Eagle-Block2A-2B-v2}; these may be the same model under different names.
\end{table}

Figure~\ref{fig:scaling} shows two patterns. First, data strategies have diverged. Between late 2023 and early 2025, several models trained on about one million episodes (Open X-Embodiment, Octo, OpenVLA, RDT-1B, SpatialVLA). Since then, frontier systems report 10,000--20,000 hours of robot data (about 10,000 for $\pi_0$~\citep{pi0} and 20,000 for LingBot-VLA~\citep{pragmaticvla}); $\pi^*_{0.6}$ states only ``tens of thousands of hours''~\citep{pi06starrecap}. At the same time, data has broadened beyond robots: recent releases add large-scale human video, such as the 20k hours of egocentric video in GR00T~N1.7 (Table~\ref{tab:groot}). Others use far less: SmolVLA trains on 22,900 community episodes~\citep{smolvla}, and TinyVLA, VLA-Adapter and VLA-0 skip robot pretraining entirely~\citep{tinyvla,vlaadapter,vla0}. Their strongest results, however, come from simulation benchmarks such as LIBERO, which test much less than real-world use (Section~\ref{sec:benchmarks}). Second, model size has not simply fallen. RT-2 (55B) is still the largest robot policy here; PaLM-E (562B) is larger but acts as a planner, not a policy. Most 2025--2026 models are 3--7B, and a new group under one billion has appeared: SmolVLA (0.45B), VLA-Adapter and SimVLA (0.5B)~\citep{simvla}, and X-VLA (0.9B)~\citep{xvla}. SmolVLA is designed to run on consumer GPUs or even CPUs, and VLA-Adapter trains in 8 hours on a single consumer-grade GPU.

{\small
\setlength{\tabcolsep}{4pt}
\begin{longtable}{L{2.45cm}L{1.25cm}L{4.9cm}L{5.55cm}}
\caption{Manipulation VLAs and precursors. OXE: Open X-Embodiment; FT: fine-tuning. $^\ast$Preprint.}\label{tab:manip}\\
\toprule
Model & Released & Training data & Key idea \\
\midrule
\endfirsthead
\toprule
Model & Released & Training data & Key idea \\
\midrule
\endhead
\bottomrule
\endfoot
RT-1~\citep{rt1} & Dec 2022 & $\sim$130k episodes, 700+ tasks, 13 robots of one type & Transformer over image and language tokens; data diversity beats volume. \\
PaLM-E~\citep{palme} & Mar 2023 & Web VL data + task-and-motion-planning (TAMP) sim, Language-Table~\citep{interactivelanguage}, mobile manipulation & 562B embodied LM as planner; positive transfer from task mixtures. \\
Diffusion Policy~\citep{diffusionpolicy} & Mar 2023 & Per-task demos (90--210 real) & Policy as conditional denoising over action sequences. \\
ACT~\citep{actaloha} & Apr 2023 & 50 demos per task on ALOHA & Action chunking with a conditional variational autoencoder (CVAE) transformer; low-cost bimanual hardware. \\
RT-2~\citep{rt2} & Jul 2023 & RT-1 data co-fine-tuned with web VQA & Coined ``VLA''; actions as text tokens; emergent semantic reasoning. \\
Open X-Embodi\-ment~\citep{openxembodiment} & Oct 2023 & Dataset: 1M+ trajectories, 22 embodiments & Standard cross-lab format; RT-X shows positive transfer. \\
Robo\-Flamingo~\citep{roboflamingo} & Nov 2023 & CALVIN language-annotated demos ($\sim$24k) & Open VLM plus explicit history policy head; light fine-tuning. \\
3D-VLA~\citep{threedvla} & Mar 2024 & 3D instruction set from robot data, RLBench, CALVIN, human videos & 3D-LLM with generated goal images and point clouds. \\
Octo~\citep{octo} & May 2024 & 800k episodes, 25 OXE datasets & Open 93M diffusion policy; adapts to new sensors and action spaces. \\
OpenVLA~\citep{openvla} & Jun 2024 & 970k curated OXE episodes & Open 7B VLA; beats RT-2-X (55B); low-rank (LoRA) fine-tuning and quantisation. \\
ECoT~\citep{ecot} & Jul 2024 & BridgeData V2~\citep{bridgedatav2} (60k) with synthetic reasoning & Embodied chain-of-thought before acting; +28 points over OpenVLA. \\
TinyVLA~\citep{tinyvla} & Sep 2024 & No robot pretraining; $\sim$50 demos per task & Compact VLM with diffusion decoder; data-efficient. \\
GR-2\npr~\citep{gr2} & Oct 2024 & 38M web video clips, then $\sim$40k robot trajectories & Video-generation pretraining, then joint video and action prediction. \\
RDT-1B~\citep{rdt1b} & Oct 2024 & 46 datasets, 1M+ trajectories; 6k bimanual FT & 1.2B diffusion transformer; unified physical action space. \\
LAPA~\citep{lapa} & Oct 2024 & Action-free video (Bridge, OXE or SSv2) & Latent actions from a vector-quantised autoencoder (VQ-VAE) enable pretraining without action labels. \\
$\pi_0$~\citep{pi0} & Oct 2024 & $\sim$10k\,h in-house (7 robot configs) + OXE & VLM plus flow-matching action expert; dexterous multi-robot skills. \\
CogACT\npr~\citep{cogact} & Nov 2024 & OXE ($\sim$1M trajectories); 400 real FT demos & VLM cognition feature conditions a diffusion action transformer. \\
FAST~\citep{fast} & Jan 2025 & DROID~\citep{droid} (75k); $\pi$ 10k\,h mixture & DCT-compressed action tokens; $5\times$ faster autoregressive training. \\
SpatialVLA~\citep{spatialvla} & Jan 2025 & 1.1M episodes, OXE subset + RH20T~\citep{rh20t} & Ego3D position encoding; adaptive action grids. \\
HAMSTER~\citep{hamster} & Feb 2025 & Off-domain: 10k Bridge + $\sim$45k DROID paths, RLBench, 770k points & High-level VLM draws 2D paths for a 3D low-level policy. \\
DexVLA~\citep{dexvla} & Feb 2025 & $\sim$100\,h cross-embodiment demos & 1B plug-in diffusion expert; embodiment curriculum. \\
Hi Robot~\citep{hirobot} & Feb 2025 & Teleop skills + VLM-generated synthetic prompts & High-level VLM handles open-ended prompts and live corrections. \\
OpenVLA-OFT~\citep{openvlaoft} & Feb 2025 & FT on LIBERO, ALOHA, BridgeData V2 & Parallel decoding, chunking, continuous actions, L1 loss. \\
GR00T N1\npr~\citep{gr00tn1} & Mar 2025 & Human video, 780k sim, 300k neural trajectories, real humanoid & Open dual-system humanoid model; generated training trajectories. \\
Gemini Robotics\npr~\citep{geminirobotics} & Mar 2025 & Thousands of hours of ALOHA 2 teleop + Gemini data & VLA plus Gemini Robotics-ER for embodied reasoning. \\
$\pi_{0.5}$~\citep{pi05} & Apr 2025 & Co-training: $\sim$100 homes, cross-embodiment, web, subtasks & Co-training enables long-horizon cleaning in unseen homes. \\
UniVLA~\citep{univla} & May 2025 & OXE subset + GNM~\citep{gnm} navigation + Ego4D & Task-centric latent actions in DINOv2~\citep{dinov2} space. \\
$\pi_{0.5}$-KI~\citep{pi05knowledgeinsulation} & May 2025 & $\pi$ mixture + VLM data & Knowledge insulation: stop gradients from the action expert. \\
SmolVLA\npr~\citep{smolvla} & Jun 2025 & 22.9k episodes, 481 community SO-100 datasets & $\sim$0.45B model; asynchronous inference; trains on one GPU. \\
RTC~\citep{realtimechunking} & Jun 2025 & Inference method on $\pi_{0.5}$ & Freeze committed actions, inpaint the rest; no retraining. \\
MolmoAct~\citep{molmoact} & Aug 2025 & OXE subset + 10k MolmoAct trajectories + web data & Depth tokens, editable trajectory traces, then actions; fully open. \\
VLA-Adapter~\citep{vlaadapter} & Sep 2025 & No robot pretraining; LIBERO, CALVIN demos & 0.5B backbone with Bridge Attention; 8\,h training on one GPU. \\
Gemini Robotics 1.5\npr~\citep{geminirobotics1p5} & Oct 2025 & ALOHA, bi-arm Franka, Apollo humanoid + web data & Motion Transfer across bodies; thinks before acting. \\
X-VLA~\citep{xvla} & Oct 2025 & 290k episodes: DROID, RoboMind~\citep{robomind}, AgiBot World~\citep{agibotworld} & Per-source soft prompts; 0.9B flow-matching transformer. \\
VLA-0\npr~\citep{vla0} & Oct 2025 & No robot pretraining; LIBERO, 100 real demos per task & Actions as plain text with an unmodified VLM. \\
MemER~\citep{memer} & Oct 2025 & 50 demos per task on $\pi_{0.5}$-DROID and Qwen2.5-VL-7B~\citep{qwen25vl} & High-level VLM selects keyframes as episodic memory. \\
AsyncVLA (flow matching)\npr~\citep{asyncvlaflowmatching} & Nov 2025 & OXE pretraining; LIBERO, Bridge, Fractal FT & Asynchronous flow matching with confidence-based self-correction. \\
$\pi^*_{0.6}$~\citep{pi06starrecap} & Nov 2025 & Multi-robot demos + autonomous rollouts + corrections & RL via advantage-conditioned policies (RECAP). \\
LingBot-VLA\npr~\citep{pragmaticvla} & Jan 2026 & $\sim$20k\,h teleop on 9 dual-arm embodiments & Data scaling from 3k to 20k\,h; 4 platforms $\times$ 100 tasks. \\
VLA-JEPA~\citep{vlajepa} & Feb 2026 & DROID (76k) + 220k human videos & JEPA-style latent world-model pretraining. \\
SimVLA\npr~\citep{simvla} & Feb 2026 & No robot pretraining; LIBERO, Fractal + Bridge; 500\,h Galaxea data for real robot & SmolVLM-0.5B~\citep{smolvlm} encoder + small flow-matching head; a controlled baseline. \\
$\pi_{0.7}$\npr~\citep{pi07} & Apr 2026 & Demos incl.\ failures, autonomous rollouts, human video & Steerable via metadata and subgoal images; zero-shot cross-embodiment. \\
Qwen-VLA\npr~\citep{qwenvla} & May 2026 & Real robots, egocentric human, 7.2M sim, VLN, VL data & One model for manipulation, navigation and trajectories. \\
\end{longtable}
}

\section{Navigation VLAs}
\label{sec:navigation}

\begin{figure}[t]
\centering
\begin{tikzpicture}[
  x=3.85cm, y=1cm,
  model/.style={draw=#1!70!black, fill=#1!10, rounded corners=2pt, align=center, font=\scriptsize, inner sep=3pt, minimum height=0.95cm, text width=2.15cm},
  model/.default=teal,
  arr/.style={-{Latex[length=1.8mm]}, line width=0.8pt, draw=black!55},
  rowlbl/.style={font=\scriptsize\bfseries, align=left, anchor=west, text=black!70},
  traits/.style={font=\scriptsize\itshape, anchor=west, text=black!55},
  grid/.style={draw=black!10, line width=0.5pt}]

  \foreach \yr in {23,24,25,26} {
    \draw[grid] (\yr,-8.35) -- (\yr,1.1);
    \node[font=\scriptsize\bfseries, text=black!55] at (\yr,-8.65) {20\yr};
  }

  % Row 1: visual navigation foundation models
  \node[rowlbl] at (22.8,0.85) {Visual navigation foundation models};
  \node[model] (vint)  at (23.46,0) {\textbf{ViNT}\\goal-image transformer};
  \node[model] (nomad) at (24.05,-1.25) {\textbf{NoMaD}\\diffusion; explore + reach};
  \node[model] (lelan) at (24.79,0) {\textbf{LeLaN}\\VLM-labelled video};
  \node[model] (omni)  at (25.71,-1.25) {\textbf{OmniVLA}\\pose, image or language goals};
  \node[model] (async) at (26.35,0) {\textbf{AsyncVLA}\\remote VLA + edge adapter};
  \draw[arr] (vint) -- (nomad); \draw[arr] (nomad) -- (lelan);
  \draw[arr] (lelan) -- (omni); \draw[arr] (omni) -- (async);
  \node[traits] at (22.8,-2.2) {real-robot data $\cdot$ waypoint outputs $\cdot$ edge deployment};

  % Row 2: VLN with video VLMs
  \node[rowlbl] at (22.8,-2.9) {Vision-and-language navigation with video VLMs};
  \node[model=blue] (navid)   at (24.13,-3.75) {\textbf{NaVid}\\monocular RGB video};
  \node[model=blue] (uninavid) at (25.0,-3.75) {\textbf{Uni-NaVid}\\4 nav tasks, one model};
  \node[model=blue] (navfom)  at (25.71,-5.0) {\textbf{NavFoM}\\quadruped, drone, car};
  \node[model=blue] (stream)  at (25.0,-5.0) {\textbf{StreamVLN}\\slow-fast memory};
  \node[model=blue] (vling)   at (26.35,-3.75) {\textbf{VLingNav}\\adaptive reasoning + RL};
  \draw[arr] (navid) -- (uninavid); \draw[arr] (uninavid) -- (navfom);
  \node[traits] at (22.8,-5.95) {Habitat simulation data $\cdot$ next-step actions $\cdot$ R2R-CE / RxR-CE};

  % Row 3: hierarchical systems
  \node[rowlbl] at (22.8,-6.65) {Hierarchical systems};
  \node[model=orange, text width=2.9cm] (mob) at (24.35,-7.5) {\textbf{Mobility VLA}\\long-context VLM + topological graph};
  \node[model=orange, text width=2.9cm] (navila) at (25.35,-7.5) {\textbf{NaVILA}\\language commands + RL locomotion};
\end{tikzpicture}
\caption{The two main lineages of navigation VLAs, with hierarchical systems in the bottom row. Arrows link works from the same group that build directly on the earlier model or its data: ViNT to AsyncVLA from the Berkeley visual-navigation line~\citep{vint,nomad,lelan,omnivlanav,asyncvlanav}, and NaVid to NavFoM from the video-VLM line~\citep{navid,uninavid,navfom}. StreamVLN~\citep{streamvln} and VLingNav~\citep{vlingnav} are independent works in the second line. The bottom row shows hierarchical systems~\citep{mobilityvla,navila}. Within each row, boxes are arranged in order of first arXiv release.}
\label{fig:navlineage}
\end{figure}
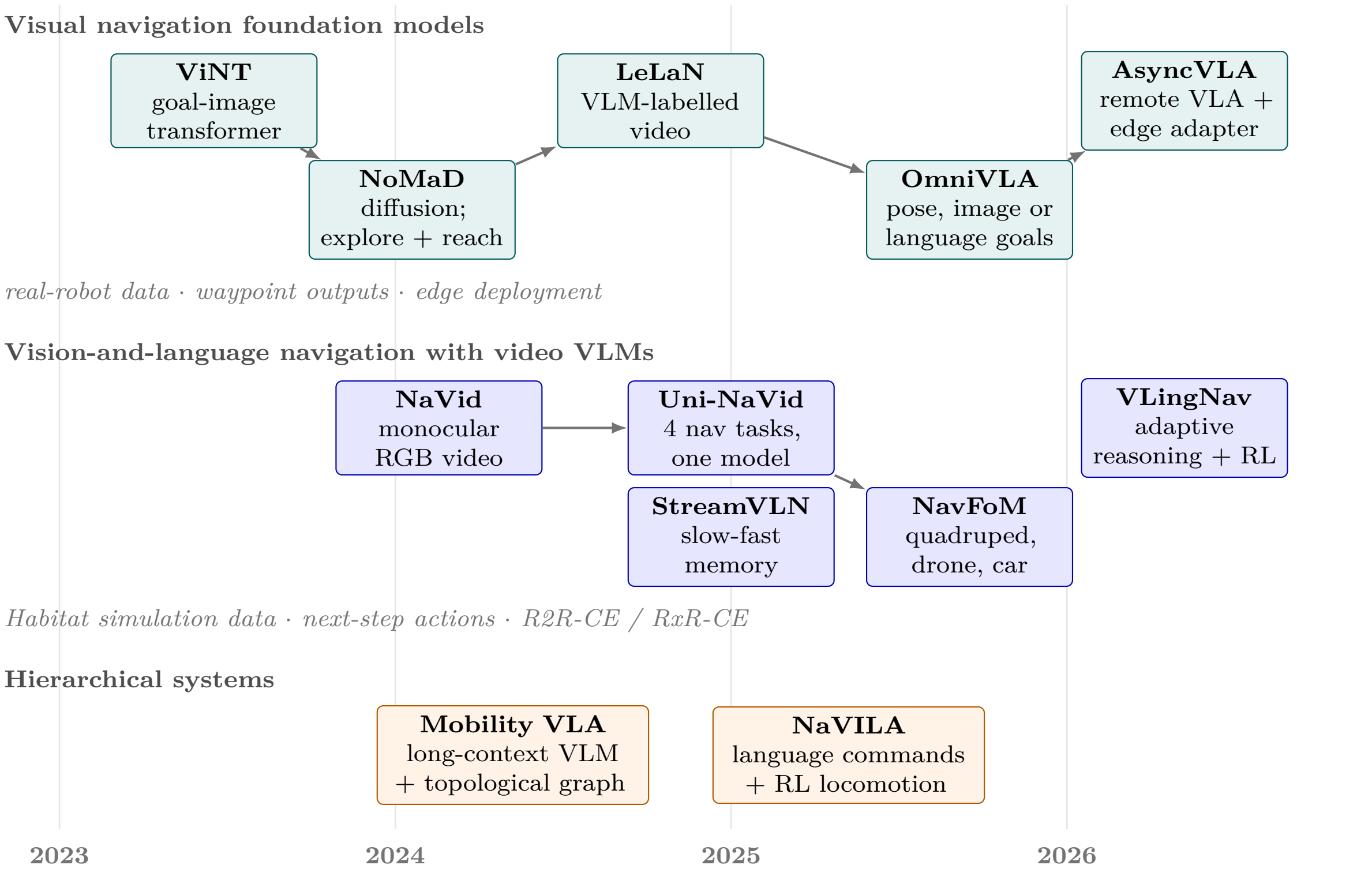

Most navigation VLAs come from two lineages with different data, outputs and evaluation; a smaller third group places a VLM over a separately trained controller (Figure~\ref{fig:navlineage}, Table~\ref{tab:nav}).

\textbf{Visual navigation foundation models.} ViNT~\citep{vint} trained a navigation transformer on more than 100 hours of real trajectories from 8 robot platforms and adapted to new goal types by replacing its goal encoder. NoMaD~\citep{nomad} used goal masking so that one diffusion policy handles both exploration and goal reaching. LeLaN~\citep{lelan} labelled about 130 hours of action-free egocentric video, including YouTube tours and human walking footage, using VLMs. OmniVLA~\citep{omnivlanav} trained on about 10k hours of navigation data with goals given as 2D poses, images, language, or combinations. AsyncVLA (navigation)~\citep{asyncvlanav} split a remote VLA from an onboard edge adapter. Models in this line output waypoints, train mostly on real robot data and target edge deployment.

\textbf{Vision-and-language navigation (VLN) with video VLMs.} NaVid~\citep{navid} showed that a video VLM can navigate from a single RGB camera's video stream, with no map, odometry or depth sensor. It trained on 510k VLN-CE samples and 763k web samples. Uni-NaVid~\citep{uninavid} unified instruction following, object search, embodied question answering and person following using 3.6M simulated samples. StreamVLN~\citep{streamvln} introduced slow-fast context modelling for streaming input. NavFoM~\citep{navfom} trained across quadrupeds, drones, wheeled robots and cars on 8.02M navigation samples. VLingNav~\citep{vlingnav} added adaptive reasoning, linguistic memory and online RL. This line trains mostly in Habitat simulation~\citep{habitat}, outputs next-step actions, and is evaluated on R2R-CE and RxR-CE~\citep{vlnce}.

\textbf{Hierarchical systems.} A smaller group keeps the VLM out of the control loop. NaVILA~\citep{navila} brought VLN to legged robots by emitting mid-level language commands that an RL locomotion policy executes. Mobility VLA~\citep{mobilityvla} needed no policy training: a long-context VLM locates the goal frame in a demonstration-tour video, and a topological-graph controller drives to it.

{\small
\setlength{\tabcolsep}{4pt}
\begin{longtable}{L{2.45cm}L{1.25cm}L{4.9cm}L{5.55cm}}
\caption{Navigation VLAs. $^\ast$Preprint.}\label{tab:nav}\\
\toprule
Model & Released & Training data & Key idea \\
\midrule
\endfirsthead
\toprule
Model & Released & Training data & Key idea \\
\midrule
\endhead
\bottomrule
\endfoot
ViNT~\citep{vint} & Jun 2023 & 100+\,h real trajectories, 8 platforms & Navigation foundation transformer; swappable goal encoders. \\
NoMaD~\citep{nomad} & Oct 2023 & GNM + SACSoN~\citep{sacson} (100+\,h) & Goal masking unifies exploration and goal reaching. \\
NaVid~\citep{navid} & Feb 2024 & 510k VLN-CE samples + 763k web data & Monocular RGB video only; history as memory. \\
Mobility VLA~\citep{mobilityvla} & Jul 2024 & No policy training; demonstration-tour video & Long-context VLM finds the goal frame; topological graph executes. \\
LeLaN~\citep{lelan} & Oct 2024 & $\sim$130\,h action-free video (robot, YouTube, walking) & VLM labelling of in-the-wild video for object navigation. \\
NaVILA~\citep{navila} & Dec 2024 & R2R-CE, RxR-CE, touring videos, VQA & Mid-level language commands executed by RL locomotion. \\
Uni-NaVid~\citep{uninavid} & Dec 2024 & 3.6M Habitat samples across 4 tasks & One video VLA for four navigation task types. \\
StreamVLN~\citep{streamvln} & Jul 2025 & 450k MP3D~\citep{matterport3d} + 300k ScaleVLN~\citep{scalevln} + 240k DAgger (expert-corrected rollouts) + 248k video QA & Slow-fast context with KV-cache reuse. \\
NavFoM~\citep{navfom} & Sep 2025 & 8.02M navigation + 4.76M QA samples & Cross-embodiment, cross-task navigation foundation model. \\
OmniVLA~\citep{omnivlanav} & Sep 2025 & $\sim$10k\,h: GNM, LeLaN, Frodobots-2K~\citep{frodobots2k}, BDD-V~\citep{bddv} & Omni-modal goals (pose, image, language). \\
VLingNav~\citep{vlingnav} & Jan 2026 & 4.5M samples incl.\ Nav-AdaCoT-2.9M; online RL & Adaptive chain-of-thought and linguistic memory. \\
AsyncVLA (navi\-gation)\npr~\citep{asyncvlanav} & Feb 2026 & GNM + SACSoN + LeLaN & Remote VLA plus onboard edge adapter; robust to 6\,s delay. \\
\end{longtable}
}

\section{Locomotion and Humanoid Whole-Body Control}
\label{sec:humanoid}

Manipulation VLAs usually output end-effector poses, and navigation VLAs output velocities or waypoints; in both cases a separate controller executes them. On a humanoid this split is harder, because the controller must keep the robot balanced while the legs, waist and arms move together, and LeVERB argues that fixed action interfaces of this kind restrict humanoids to slow, quasi-static tasks~\citep{leverb}. Suitable data is also scarce: humanoid teleoperation is expensive~\citep{wholebodyvla}, and dynamically consistent demonstrations are rare~\citep{wolfvla}. Two designs have emerged (Table~\ref{tab:humanoid}).

\paragraph{Hierarchical: a vision-language layer over a learned whole-body controller.} This mirrors the hierarchical VLAs of Section~\ref{sec:taxonomy}, and NaVILA's language commands over an RL locomotion policy are a navigation precursor~\citep{navila}. Humanoid-VLA~\citep{humanoidvla} pre-aligns language with motion using human motion datasets paired with text, then adds egocentric video through parameter-efficient fine-tuning, on top of a whole-body control architecture. LeVERB~\citep{leverb} trains a vision-language policy to emit latent ``verbs'' that a reinforcement-learned whole-body controller turns into dynamics-level commands. It also introduces a closed-loop benchmark of over 150 tasks in 10 categories, where it reaches 58.5\% overall success, 7.8 times a naive hierarchical baseline. SONIC~\citep{sonic} scales the controller itself: a motion-tracking foundation model of up to 42M parameters, trained on 700 hours of motion capture with 21k GPU hours, whose token space serves both VR teleoperation and VLAs. WholeBodyVLA~\citep{wholebodyvla} learns latent actions from action-free egocentric human video and pairs the VLA with an RL policy specialised for loco-manipulation moves such as advancing, turning and squatting. On the AgiBot X2 humanoid it reaches 78.0\% average success, 21.3 percentage points above the strongest VLA baseline (OpenVLA-OFT, 56.7\%) and 14 points above a modular, non-VLA baseline (64.0\%).

\paragraph{End-to-end: language to joint actions.} SENTINEL~\citep{sentinel} maps language and proprioception directly to low-level actions with flow matching, with no intermediate representation, and refines them with a residual RL head on a Unitree G1. Its training data comes from human motion datasets tracked in simulation by a pretrained whole-body controller. WOLF-VLA~\citep{wolfvla} replaces human motion with optimal control, synthesising 277 hours of dynamically feasible trajectories across six locomotion task families for the RH5 humanoid. Unlike these two, HAF~\citep{haf} does not train a new model. It adapts an existing flow-matching VLA to generate leg, waist and arm actions in turn, then improves it with reinforcement learning that adjusts only the noise fed into the denoising process, leaving the VLA's weights unchanged.

\paragraph{Patterns.} Three themes from manipulation reappear. Humanoid VLAs lean on non-robot data, from human motion capture and egocentric video to simulation and optimal control, because robot teleoperation is scarce. Reinforcement learning returns as a refinement stage (SENTINEL's residual head, HAF's latent RL). And industrial models are moving the same way: GR00T~N1.6 added Unitree G1 loco-manipulation data, and the N1.7 release documents SONIC-based whole-body control~\citep{grootn16readme,grootn17readme}.

{\small
\setlength{\tabcolsep}{4pt}
\begin{longtable}{L{2.45cm}L{1.25cm}L{4.9cm}L{5.55cm}}
\caption{Humanoid whole-body control with vision and language. $^\ast$Preprint.}\label{tab:humanoid}\\
\toprule
Model & Released & Training data & Key idea \\
\midrule
\endfirsthead
\toprule
Model & Released & Training data & Key idea \\
\midrule
\endhead
\bottomrule
\endfoot
Humanoid-\allowbreak VLA\npr~\citep{humanoidvla} & Feb 2025 & Human motion datasets with text, egocentric video, self-generated question--answer pairs from motion & Language--motion pre-alignment, then video-conditioned fine-tuning over whole-body control. \\
LeVERB\npr~\citep{leverb} & Jun 2025 & Synthetic: 154 trajectories $\times$ 100 randomisations, 17.1\,h of rendered rollouts plus 2.7\,h language-only & Latent ``verbs'' from a vision-language policy drive an RL whole-body controller; 150-task benchmark. \\
SONIC~\citep{sonic} & Nov 2025 & 700\,h of motion capture (100M+ frames), 21k GPU hours & Motion-tracking controller up to 42M parameters; one token space for VR teleoperation and VLAs. \\
SENTINEL~\citep{sentinel} & Nov 2025 & Human motion tracked in simulation by a pretrained controller, with text & End-to-end language and proprioception to joint actions by flow matching; residual RL head; Unitree G1. \\
Whole\-Body\-VLA~\citep{wholebodyvla} & Dec 2025 & Action-free egocentric human video (latent actions) plus AgiBot X2 teleoperation & Latent VLA plus a loco-manipulation RL policy; +21.3 points over OpenVLA-OFT. \\
WOLF-VLA\npr~\citep{wolfvla} & Jun 2026 & 277\,h of optimal-control trajectories, six locomotion task families & VLA trained on dynamically feasible trajectories with egocentric vision and language; RH5 humanoid. \\
HAF\npr~\citep{haf} & Aug 2026 & Demonstrations for seven real-world loco-manipulation tasks, then online RL & Three-stage action flow on a pretrained flow-matching VLA; RL in a compact noise subspace. \\
\end{longtable}
}

\section{Convergence of Manipulation and Navigation}
\label{sec:convergence}

The two communities have long used different action spaces (end-effector or joint commands versus waypoints), benchmarks (LIBERO, SimplerEnv~\citep{simplerenv} and CALVIN versus VLN-CE) and model families. Two recent models are early signs of unification:
\begin{itemize}
  \item UniVLA~\citep{univla} learns latent actions from manipulation data, navigation data and human video together, and reports results on both manipulation benchmarks and R2R.
  \item Qwen-VLA~\citep{qwenvla} treats manipulation, navigation and trajectory prediction as one prediction problem. A single model reaches 97.9\% on LIBERO and 59.6\% success on RxR. Its authors find no significant loss from training on several robot types together, in an ablation on two benchmarks.
\end{itemize}
Two related lines stop short of unification. Mobile manipulators such as $\pi_{0.5}$~\citep{pi05} combine moving and manipulating on one robot, but are trained and evaluated as manipulation systems. NavFoM~\citep{navfom} covers very different bodies, from quadrupeds to cars, but only for navigation. In our view, the main obstacle to a single model for both skills may be shifting from architecture to data and evaluation. In terms of the life-cycle axes, both unifying models act at the training stage, through pooled data and a shared action interface (latent actions in UniVLA, one prediction problem in Qwen-VLA), while their evaluation remains split across separate manipulation and navigation benchmarks.

\section{VLAs and World-Action Models}
\label{sec:wam}

A second family of robot foundation models starts from video rather than from language. A \emph{world model} predicts how the environment changes as an agent acts~\citep{ha2018worldmodels,lecun2022path}. \emph{World-action models} (WAMs) extend this idea by predicting future observations and robot actions together. The earliest robotics use of the term we found is DyWA~\citep{dywa}; most current WAMs start from a model pretrained to generate video. Where a VLA learns $\pi(A_t \mid o_t)$ from action supervision alone, a WAM learns to predict future observations together with actions,
\begin{equation}
  p\big(O_{t+1:t+k},\, A_t \mid o_t, \ell_t\big),
\end{equation}
where $O_{t+1:t+k}$ are future frames, or latent embeddings of them. World-action models~\citep{surveywam} and world models for robot learning~\citep{surveyworldmodelrobot} have dedicated surveys; here we compare WAMs with VLAs. The two families start from different pretrained models and fail in different ways (Figure~\ref{fig:vlawam}, Table~\ref{tab:wam}).

\begin{figure}[t]
\centering
\begin{tikzpicture}[
  font=\small,
  box/.style={draw, rounded corners=2pt, align=center, minimum height=0.95cm, inner sep=4pt, text width=3.6cm},
  vlm/.style={box, fill=blue!8},
  vid/.style={box, fill=orange!12},
  outbox/.style={box, fill=black!4, text width=2.3cm},
  io/.style={align=center, font=\footnotesize},
  loss/.style={font=\footnotesize\itshape, text=red!55!black},
  arr/.style={-{Latex[length=2mm]}, thick}]

  % VLA
  \node[font=\bfseries\small] at (0,1.15) {(a) Vision-language-action model};
  \node[io] (vin) at (0,0.45) {images + instruction};
  \node[vlm, below=0.4cm of vin] (vlm) {VLM backbone\\[-1pt]{\footnotesize prior: image--text data}};
  \node[box, fill=blue!3, below=0.4cm of vlm] (ahead) {action head};
  \node[outbox, below=0.4cm of ahead] (vact) {action chunk};
  \draw[arr] (vin) -- (vlm); \draw[arr] (vlm) -- (ahead); \draw[arr] (ahead) -- (vact);
  \node[loss, right=0.15cm of vact, align=left] {action loss};

  % WAM
  \node[font=\bfseries\small] at (7.6,1.15) {(b) World-action model};
  \node[io] (win) at (7.6,0.45) {images + instruction};
  \node[vid, below=0.4cm of win, text width=4.2cm] (vid) {video generation backbone\\[-1pt]{\footnotesize prior: video data}};
  \node[box, fill=orange!4, below=0.4cm of vid, xshift=-1.55cm, text width=2.5cm] (frames) {future frames\\or latents};
  \node[box, fill=orange!4, below=0.4cm of vid, xshift=1.55cm, text width=2.5cm] (adec) {action decoder};
  \node[outbox, below=0.4cm of adec] (wact) {action chunk};
  \draw[arr] (win) -- (vid);
  \draw[arr] (vid.south) ++(-0.4,0) -- (frames.north);
  \draw[arr] (vid.south) ++(0.4,0) -- (adec.north);
  \draw[arr, dashed] (frames.east) -- (adec.west);
  \draw[arr] (adec) -- (wact);
  \node[loss, below=0.12cm of frames, align=center] {frame loss};
  \node[loss, right=0.15cm of wact, align=left] {action loss};
  \node[font=\footnotesize, text=black!60, align=center, below=0.3cm of wact, xshift=-1.55cm] {optional uses of predicted futures: planning, data generation, evaluation};
\end{tikzpicture}
\caption{Where the two families start. (a) A VLA adapts a vision-language model and is supervised only on actions. (b) A world-action model adapts a video model and is supervised on future observations as well as actions; predicted futures can inform the action decoder (dashed) and support planning, synthetic data or evaluation. Table~\ref{tab:wam} compares them in detail.}
\label{fig:vlawam}
\end{figure}
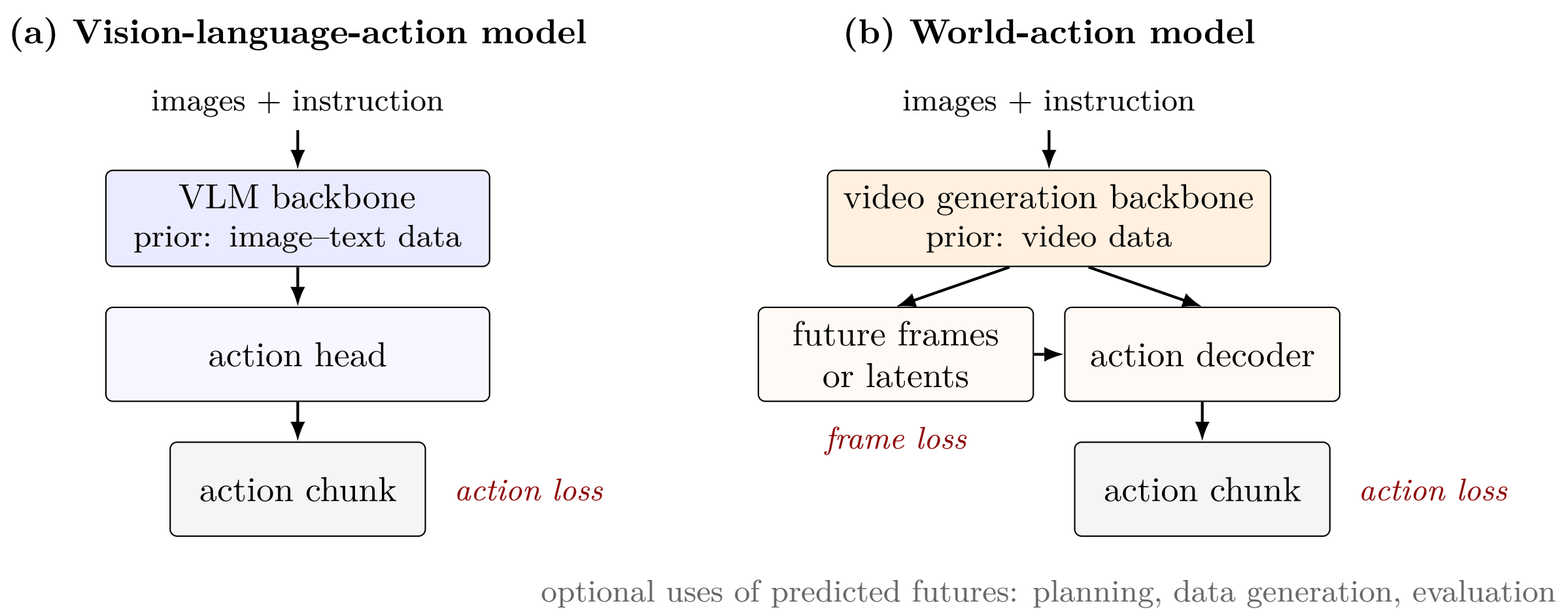 As the UniWAM authors argue~\citep{uniwam}, a VLA's vision-language model has learned language and object meaning from image--text data, but its only training signal about physics is robot actions. A WAM's video model has learned how scenes change over time, but not how to follow or reason about language.

\begin{table}[!htb]
\centering\small
\caption{VLAs and world-action models compared. $^\ast$Preprint.}
\label{tab:wam}
\begin{tabularx}{\linewidth}{L{2.6cm}XX}
\toprule
 & Vision-language-action model (VLA) & World-action model (WAM) \\
\midrule
Pretrained prior & Vision-language model trained on image--text data & Video generation model trained on video \\
Training signal & Robot actions, often co-trained with VQA and web data~\citep{rt2,pi05} & Future frames or latents together with actions~\citep{uva,worldvla} \\
Predicts & Action chunk (sometimes text subtasks or reasoning~\citep{pi05,ecot}) & Future observations and an action chunk; some also values~\citep{cosmospolicy} \\
Strengths & Language following, semantic generalisation, reasoning~\citep{rt2} & Physical dynamics, temporal consistency, use of action-free video~\citep{unipi,dreamgen}, test-time planning~\citep{cosmospolicy} \\
Weaknesses & Weak grounding in dynamics~\citep{uniwam}; relies on action-labelled data & Weaker semantics and reasoning~\citep{uniwam}; costly video backbones~\citep{lewam} \\
Examples & RT-2~\citep{rt2}, OpenVLA~\citep{openvla}, $\pi_0$~\citep{pi0}, $\pi_{0.5}$~\citep{pi05}, GR00T~N1\npr~\citep{gr00tn1} & UniPi~\citep{unipi}, VPP~\citep{vpp}, UVA~\citep{uva}, WorldVLA\npr~\citep{worldvla}, Genie Envisioner~\citep{genieenvisioner}, Cosmos Policy~\citep{cosmospolicy}, DreamZero\npr~\citep{dreamzero}, LeWAM\npr~\citep{lewam} \\
\bottomrule
\end{tabularx}
\end{table}

\paragraph{From video plans to joint models.} Two terms recur here: a \emph{forward dynamics} model predicts the next observation from the current one and an action, and an \emph{inverse dynamics} model infers the action that links two observations. UniPi~\citep{unipi} cast decision-making as text-conditioned video generation: a model synthesises a video of the planned behaviour, and an inverse dynamics model recovers actions from the generated frames. Video Prediction Policy~\citep{vpp} skips full video generation: it feeds a video diffusion model's internal features about what will happen next to an inverse dynamics model, giving an 18.6\% relative improvement on CALVIN ABC$\rightarrow$D. The Unified Video Action model~\citep{uva} learns one shared representation of video and actions, with separate diffusion heads for each. Action inference can then skip video generation entirely, and masking different inputs lets the same model act as a forward or an inverse dynamics model. WorldVLA~\citep{worldvla} merges a VLA and a world model in one autoregressive model and reports that each improves the other.

\paragraph{Video foundation models as policy backbones.} Genie Envisioner~\citep{genieenvisioner} builds a robotics platform around an instruction-conditioned video diffusion model, with a flow-matching action decoder, an action-conditioned neural simulator and a benchmark for world-model quality. Cosmos Policy~\citep{cosmospolicy} turns a pretrained video model into a policy by fine-tuning alone, keeping its architecture: actions, future states and value estimates are written into the model's latent frames. Because it predicts future states and values, it can plan at test time; it reports 98.5\% on LIBERO and 67.1\% on RoboCasa~\citep{robocasa}. DreamZero~\citep{dreamzero} builds a 14B world-action model on a pretrained image-to-video diffusion model, predicting future video and actions jointly with an autoregressive transformer trained by flow matching. After each action chunk it replaces its predicted frames with the real observations, so errors do not accumulate, and model and system optimisations give a 38$\times$ speed-up that allows closed-loop control at about 7\,Hz. Trained on about 500 hours of diverse, non-repetitive real-robot data, it reports more than twice the average task progress of state-of-the-art VLAs on unseen tasks and environments, and it adapts to a new robot from 30 minutes of play data.

\paragraph{Beyond video diffusion.} Two 2026 preprints take WAMs in different directions. LeWAM~\citep{lewam} makes them cheaper: it replaces the video diffusion backbone with predictive JEPA embeddings and reaches 92.3\% on RoboTwin~2.0~\citep{robotwin2} with 0.4B trainable parameters. UniWAM~\citep{uniwam} moves them towards VLAs by adding a reasoning module, the ``physical reasoner'', alongside its video generator and action predictor in one architecture. It reports a log-linear scaling law for co-training on human and robot data.

\paragraph{Three roles for world models.} World models now serve robot learning in three ways:
\begin{itemize}
  \item \textbf{As the policy backbone}, as in the WAMs above.
  \item \textbf{As a data generator.} DreamGen~\citep{dreamgen} turns video world models into a source of synthetic ``neural trajectories''. With them, a humanoid learned 22 new behaviours from teleoperation data for a single task, and DreamGen's video benchmark correlates with downstream policy success. GR00T~N1 used about 300k such trajectories~\citep{gr00tn1}.
  \item \textbf{As an evaluator.} Neural simulators such as Genie Envisioner's GE-Sim produce closed-loop rollouts for policy development~\citep{genieenvisioner}.
\end{itemize}

\paragraph{A blurring boundary.} VLAs increasingly predict the future: 3D-VLA, GR-2, VLA-JEPA and $\pi_{0.7}$ all generate goal images, video or future latents (Section~\ref{sec:taxonomy}). WAMs, in turn, are adding VLM-style reasoning, as UniWAM does. A useful working distinction is that VLAs are \emph{understanding-first} and WAMs are \emph{dynamics-first}; the strongest systems are likely to need both.

\section{Benchmarks and Fair Evaluation}
\label{sec:benchmarks}

\paragraph{Is there a standard benchmark?} Not yet. No single benchmark is accepted as the measure of a VLA. Instead, a small set of simulation suites has become the de facto standard (Table~\ref{tab:bench}), while real-world evaluation remains lab-specific and rarely comparable across papers.

\begin{table}[t]
\centering\small
\caption{Benchmarks most often used to evaluate VLAs. $^\ast$Preprint.}
\label{tab:bench}
\begin{tabularx}{\linewidth}{L{2.6cm}L{2.5cm}X}
\toprule
Benchmark & Venue & What it tests \\
\midrule
LIBERO~\citep{libero} & NeurIPS 2023 (D\&B) & Simulated Franka tasks in suites for spatial layout, objects, goals and long horizons; the most widely reported VLA benchmark. \\
CALVIN~\citep{calvin} & RA-L 2022 & Long chains of language-conditioned tasks across four simulated environments; generalisation from environments A--C to D. \\
SimplerEnv~\citep{simplerenv} & CoRL 2024 & Simulated replicas of the Google Robot and WidowX setups, built to correlate with real-world results. \\
RoboCasa~\citep{robocasa} & RSS 2024 & 100 household kitchen tasks in large-scale simulation with generated assets. \\
VLABench~\citep{vlabench} & ICCV 2025 & 100 task categories and over 2,000 objects testing world knowledge, implicit instructions and long-horizon reasoning. \\
RoboTwin 2.0~\citep{robotwin2} & ICML 2026 & 50 dual-arm tasks on five embodiments with domain randomisation over clutter, lighting, background, table height and language. \\
R2R-CE / RxR-CE~\citep{vlnce} & ECCV 2020 & Vision-and-language navigation in continuous Habitat environments, without navigation-graph assumptions. \\
\bottomrule
\end{tabularx}
\end{table}

\paragraph{What goes wrong with current practice.}
\begin{itemize}
  \item \textbf{Saturation.} LIBERO scores now cluster near the ceiling: OpenVLA-OFT reports 97.1\%~\citep{openvlaoft}, Qwen-VLA 97.9\%~\citep{qwenvla}, X-VLA about 98\%~\citep{xvla} and SimVLA 98.6\%~\citep{simvla}. VLA-0, VLA-Adapter and SimVLA reach these scores without any robot pretraining~\citep{vla0,vlaadapter}, so the benchmark can no longer show what large-scale pretraining adds.
  \item \textbf{Memorisation and brittleness.} Two independent perturbation suites agree that high LIBERO scores overstate robustness. In LIBERO-PRO~\citep{liberopro}, OpenVLA and $\pi_0$, both above 90\% on standard LIBERO, drop to 0\% once objects are moved slightly from their training positions; $\pi_{0.5}$ holds up better, keeping 38\% on one task suite. LIBERO-Plus~\citep{liberoplus} tests models including OpenVLA-OFT and $\pi_0$ under seven kinds of perturbation: changing the camera viewpoint or the robot's starting pose cuts success from about 95\% to under 30\%, and many models largely ignore the instruction.
  \item \textbf{Confounded comparisons.} Architectural changes usually arrive together with different data, backbones and training schedules, so many reported gains are hard to attribute; SimVLA's controlled baseline (Section~\ref{sec:manipulation}) makes this concrete~\citep{simvla}.
\end{itemize}

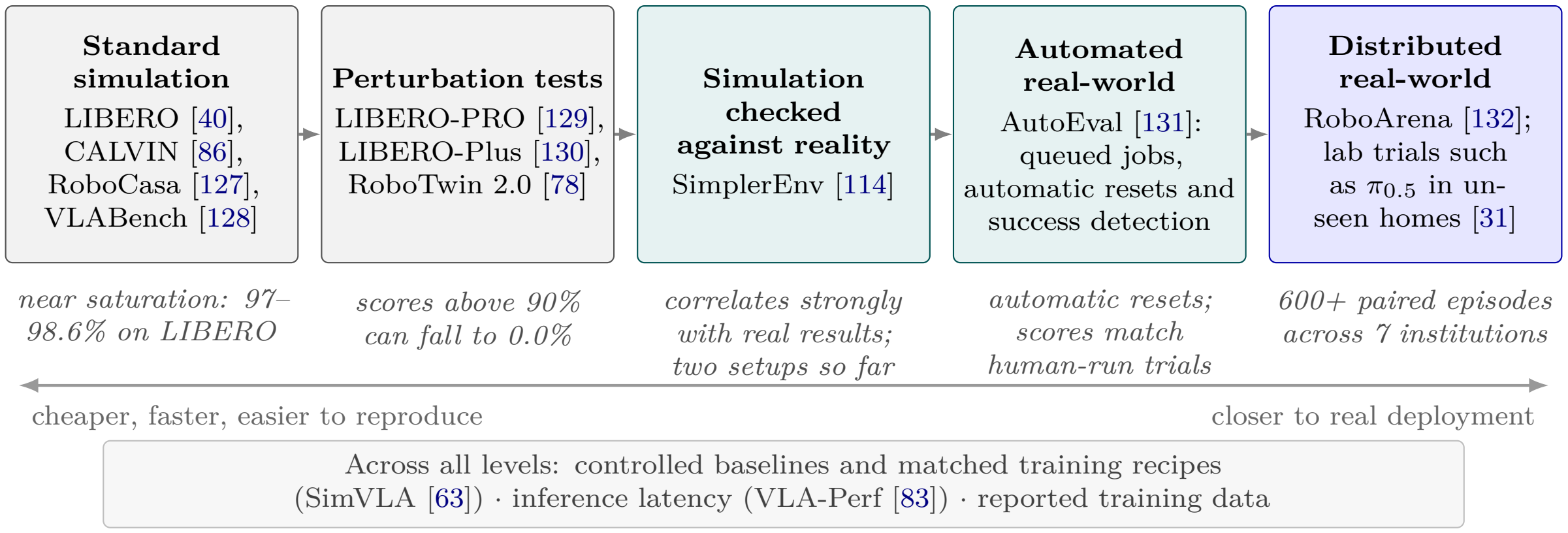
\begin{figure}[t]
\centering
\begin{tikzpicture}[
  tier/.style={draw=#1!65!black, fill=#1!10, rounded corners=2pt, align=center, font=\scriptsize, text width=2.75cm, minimum height=2.6cm, inner sep=3pt, anchor=north},
  finding/.style={font=\scriptsize\itshape, align=center, text width=2.85cm, text=black!70, anchor=north},
  arr/.style={-{Latex[length=1.8mm]}, line width=0.8pt, draw=black!50}]

  \node[tier=gray] (t1) at (1.55,0) {\textbf{Standard simulation}\\[2pt] LIBERO~\citep{libero}, CALVIN~\citep{calvin}, RoboCasa~\citep{robocasa}, VLABench~\citep{vlabench}};
  \node[tier=gray] (t2) at (4.75,0) {\textbf{Perturbation tests}\\[2pt] LIBERO-PRO~\citep{liberopro}, LIBERO-Plus~\citep{liberoplus}, RoboTwin~2.0~\citep{robotwin2}};
  \node[tier=teal] (t3) at (7.95,0) {\textbf{Simulation checked\\against reality}\\[2pt] SimplerEnv~\citep{simplerenv}};
  \node[tier=teal] (t4) at (11.15,0) {\textbf{Automated real-world}\\[2pt] AutoEval~\citep{autoeval}: queued jobs, automatic resets and success detection};
  \node[tier=blue] (t5) at (14.35,0) {\textbf{Distributed real-world}\\[2pt] RoboArena~\citep{roboarena}; lab trials such as $\pi_{0.5}$ in unseen homes~\citep{pi05}};
  \foreach \a/\b in {t1/t2, t2/t3, t3/t4, t4/t5} \draw[arr] (\a.east) -- (\b.west);

  \node[finding] at (1.55,-2.75) {near saturation: 97--98.6\% on LIBERO};
  \node[finding] at (4.75,-2.75) {scores above 90\% can fall to 0.0\%};
  \node[finding] at (7.95,-2.75) {correlates strongly with real results; two setups so far};
  \node[finding] at (11.15,-2.75) {automatic resets; scores match human-run trials};
  \node[finding] at (14.35,-2.75) {600+ paired episodes across 7 institutions};

  \draw[{Latex[length=2mm]}-{Latex[length=2mm]}, line width=0.8pt, draw=black!40] (0.2,-3.85) -- (15.7,-3.85);
  \node[font=\scriptsize, text=black!60, anchor=north west] at (0.2,-3.92) {cheaper, faster, easier to reproduce};
  \node[font=\scriptsize, text=black!60, anchor=north east] at (15.7,-3.92) {closer to real deployment};

  \node[draw=black!25, rounded corners=2pt, font=\scriptsize, text=black!75, fill=black!3, inner sep=4pt, text width=13.5cm, align=center] at (7.95,-4.85) {Across all levels: controlled baselines and matched training recipes (SimVLA~\citep{simvla}) $\cdot$ inference latency (VLA-Perf~\citep{vlaperf}) $\cdot$ reported training data};
\end{tikzpicture}
\caption{Levels of VLA evaluation, ordered qualitatively from cheap and reproducible to close to real deployment, with the key finding reported at each level. Fair comparison increasingly combines several levels rather than relying on a single simulation score.}
\label{fig:evalladder}
\end{figure}

\paragraph{Efforts toward fair evaluation.} Figure~\ref{fig:evalladder} arranges these efforts from cheap and reproducible to close to real deployment.

\begin{itemize}
  \item \textbf{Robustness-oriented benchmarks.} LIBERO-PRO and LIBERO-Plus replace fixed test scenes with systematic perturbations; RoboTwin~2.0 builds domain randomisation into the benchmark itself.
  \item \textbf{Simulation validated against reality.} SimplerEnv~\citep{simplerenv} builds simulated copies of real robot setups, tuned to match their control behaviour and appearance, and finds that policy results in these copies correlate strongly with real-world results.
  \item \textbf{Scalable real-world evaluation.} AutoEval~\citep{autoeval} automates real-robot testing: it detects success and resets the scene itself, so evaluations can run day and night, and its scores closely match human-run trials; its WidowX scenes are open to the community. RoboArena~\citep{roboarena} instead spreads evaluation across seven institutions that use the same DROID robot. Evaluators compare two policies blind, and from more than 600 such comparisons it ranks policies more accurately than a single central evaluation.
  \item \textbf{World models as evaluators.} Video-model simulators are a newer option: DreamGen Bench~\citep{dreamgen} reports that its video-model scores correlate with downstream policy success, and GE-Sim~\citep{genieenvisioner} runs policies in a generated world, but neither has yet been checked against real-robot results.
  \item \textbf{Controlled baselines.} SimVLA argues for reporting results against a fixed, transparent baseline and recipe, so that gains can be attributed to the change being studied.
  \item \textbf{Efficiency as a metric.} VLA-Perf~\citep{vlaperf} provides an analytical model of inference latency across model designs and on-device, edge and cloud deployment, making speed comparable alongside success rate.
\end{itemize}
Together these efforts point to a shared protocol: perturbed rather than fixed test scenes, simulation scores validated against real robots, distributed real-world comparison, matched training recipes, and reporting of latency and training data alongside success. Appendix~\ref{app:guide} turns this protocol into a checklist for evaluating a new model.

\section{Open Challenges and Research Directions}
\label{sec:challenges}

We group the open challenges by life-cycle stage, as in Section~\ref{sec:taxonomy}, and add those that cut across it. Each ends with a research question that the evidence in this survey motivates.

\subsection{Model}

\paragraph{Memory beyond minutes.} Keyframe and slow-fast memories now support tasks that need minutes of recall~\citep{memer,streamvln}, yet only a minority of the models in Table~\ref{tab:axesgrid} keep any history, and those are mostly navigation models. \emph{Open question: can memory scale from minutes to hours, and combine a spatial map of where the robot has been with episodic recall of what it did?}

\paragraph{World models inside VLAs.} Latent and generative world models are entering VLAs~\citep{gr2,vlajepa,pi07}, and WAMs are adding VLM-style reasoning~\citep{uniwam} (Section~\ref{sec:wam}). Yet only two of the 22 models in Table~\ref{tab:axesgrid}, VLA-JEPA and $\pi_{0.7}$, predict future observations, and controlled comparisons with and without future prediction remain rare; WorldVLA's report that the two improve each other is an exception~\citep{worldvla}. \emph{Open question: for which tasks does predicting the future repay its cost in compute and latency, and how should world models used to generate data or evaluate policies be validated against real robots?}

\subsection{Training}

\paragraph{Data-mixture design.} Mixture weights are usually set by heuristics, for example per-dataset weighting by $n^{0.43}$ in $\pi_0$~\citep{pi0} or by size and diversity in Octo~\citep{octo}, together with ad hoc rules such as dropping DROID late in training~\citep{spatialvla}. The few controlled ablations show that sources differ in what they teach: in $\pi_{0.5}$, cross-embodiment data mattered most in unseen homes and web data mainly for unseen objects~\citep{pi05}. \emph{Open question: which data sources build which capabilities, and can mixtures be designed from such measurements rather than from total hours?}

\paragraph{Learning from deployment.} RECAP~\citep{pi06starrecap} and online RL in navigation~\citep{vlingnav} show that VLAs can improve from their own experience, but only seven of the 22 models in Table~\ref{tab:axesgrid} make a distinctive post-training choice, and RECAP still relies on expert interventions. \emph{Open question: can a VLA improve safely from autonomous experience across many tasks and robots, with little human supervision?}

\paragraph{Whole-body data.} Humanoid VLAs still mostly rely on a separately trained whole-body controller, and end-to-end models are trained largely on synthetic or human-motion data~\citep{sentinel,wolfvla}, because humanoid teleoperation is expensive (Section~\ref{sec:humanoid}). \emph{Open question: can optimal-control and human-motion data be combined with limited real teleoperation to give VLAs agile, contact-rich whole-body control on real humanoids?}

\subsection{Deployment}

\paragraph{Latency and edge deployment.} Asynchronous execution~\citep{smolvla,realtimechunking,asyncvlanav} hides latency, and sub-billion models run on commodity hardware. But every model in Table~\ref{tab:deploy} that names its hardware runs on a desktop or data-centre GPU, and embedded robot computers appear only for the edge part of a split system. \emph{Open question: can large VLAs be compressed, for example by quantisation, which OpenVLA shows preserves real-robot performance~\citep{openvla}, or by distillation, which the authors of VLA-0 propose~\citep{vla0}, to run on robot hardware without losing their generality?}

\subsection{Across the life cycle}

\paragraph{Unifying skills.} UniVLA and Qwen-VLA unify manipulation and navigation through pooled data and a shared action interface, but their results are still reported on separate benchmarks, and mobile manipulators are trained and evaluated as manipulation systems (Section~\ref{sec:convergence}). \emph{Open question: how should a single model that both moves and manipulates be evaluated on tasks that need both skills at once?}

\paragraph{Predictive evaluation.} Simulation is growing as a data source (GR00T~N1, Qwen-VLA), yet few works measure how well simulated success predicts real-world transfer, and LIBERO scores near saturation hide brittleness (Section~\ref{sec:benchmarks}). SimplerEnv~\citep{simplerenv} validated simulated scores against real robots for two setups. \emph{Open question: can benchmark scores be validated against real robots across many embodiments, so that simulation results predict deployment?}

\paragraph{Reporting standards.} While building Tables~\ref{tab:manip} and~\ref{tab:nav}, we found that training data is reported inconsistently:
\begin{itemize}
  \item only 16 of the 43 manipulation works give a single robot-episode count;
  \item several report only hours, or qualitative descriptions such as ``thousands of hours'' or ``tens of thousands of hours'';
  \item eleven state no parameter count for the deployed policy;
  \item inference speed, when reported, is given as throughput, latency per chunk, frames per second or effective control rate, on different GPUs (Table~\ref{tab:deploy}).
\end{itemize}
This makes like-for-like comparison of data efficiency and compute impossible. We suggest that VLA papers report, at minimum, the robot episodes and hours used in each training stage, the share of non-robot data, the number of embodiments, the total parameter count at inference, and the latency per action chunk on named hardware; Appendix~\ref{app:guide} lists these as a checklist.

\paragraph{Announcements without published methods.} Three systems announced in 2026 bear on these directions but have no published method details, so we do not review them or report their results. NVIDIA previewed GR00T~N2, a robot foundation model based on DreamZero with a world-action-model architecture, in March 2026~\citep{gr00tn2}. Google introduced Gemini~4 Argon in October 2026~\citep{geminiargon} and Gemini~4 Hydrogen in November 2026~\citep{geminihydrogen}, general Gemini models that control robot arms by planning the task and writing and running control code, rather than by outputting actions as a VLA does.

\section{Conclusion}
\label{sec:conclusion}
VLA models have advanced from a proof of concept to a diverse ecosystem in under four years. Followed through their life cycle, the main shifts are clear. In the model, manipulation VLAs now mostly predict action chunks with flow-matching or diffusion heads, navigation VLAs mostly still predict one step at a time, and a growing number make their reasoning explicit or predict future observations. In training, data has broadened from teleoperated robot data to community-collected episodes, human video, simulation and the robot's own attempts, while some small models skip robot pretraining altogether, although their strongest results are on simulation benchmarks. In deployment, models have split by size: a new group under one billion parameters runs on consumer GPUs, and asynchronous execution hides the latency of larger ones. Generalist VLAs that handle both manipulation and navigation are emerging, and world-action models offer a complementary, dynamics-first route. Evaluation is shifting from saturated simulation scores toward robustness tests and distributed real-world comparison. The most pressing directions are memory beyond minutes, principled data mixtures, evaluation that predicts real-world transfer, safe learning from deployment, and consistent reporting of data and latency. We hope the life-cycle taxonomy, the data-centred tables and the practitioner's guide in Appendix~\ref{app:guide} help researchers understand the field, build their own models and choose where to contribute.

\section*{Acknowledgements}
We thank arXiv for open access to the surveyed works, and their authors for sharing them there. Figure~\ref{fig:scaling} was made with Matplotlib~\citep{matplotlib} and Figure~\ref{fig:arch} with TikZ.

S.~Kadam and S.~Parsa developed the idea for this survey together. S.~Kadam designed the method and the organisation of the content, and directed and reviewed every stage of the work. S.~Parsa reviewed the manuscript and proposed the design-axis grid and the deployment table.

The authors designed this survey, specified the literature search, the selection criteria and the structure of the paper, and directed each step of the work. During the preparation of this work the authors used Claude Code (Anthropic), running the Claude Opus~5.5 model, as an assistant for tasks they specified: running the literature searches they defined with web search, Semantic Scholar and Crossref; extracting reported numbers from the papers; drafting text and \LaTeX{} formatting; and writing the TikZ and Matplotlib code from which the figures were drawn. Each claim in the paper was verified against the full-text PDF of the cited work. After using this tool, the authors reviewed and edited the content as needed and take full responsibility for the content of the published article.

\bibliographystyle{unsrtnat}
\bibliography{refs}

\clearpage
\appendix
\section{Building a VLA: Decisions Along the Life Cycle}
\label{app:guide}

This appendix turns the evidence of Sections~\ref{sec:taxonomy}--\ref{sec:benchmarks} into practical guidance for researchers who want to use or build a VLA. For each axis we state the decision, a default supported by the surveyed work, and when to deviate. The defaults reflect results reported up to October 2026 and should be re-checked as the field moves.

\subsection{Choosing a starting point}
Table~\ref{tab:startingpoint} matches common situations to a starting point among the surveyed works.

\begin{table}[!htb]
\centering\small
\caption{Starting points for common situations, with the evidence behind each.}
\label{tab:startingpoint}
\begin{tabularx}{\linewidth}{L{3.6cm}L{3.0cm}X}
\toprule
Situation & Start from & Why \\
\midrule
One consumer GPU and a low-cost arm & SmolVLA~\citep{smolvla} & 0.45B parameters; trains on one GPU; pretrained on community data from low-cost SO-100 arms; asynchronous inference (Sections~\ref{sec:data} and~\ref{sec:realtime}). \\
Fine-tuning an open VLA for one robot or task set & OpenVLA-OFT recipe~\citep{openvlaoft} & Parallel decoding, action chunking, continuous actions and an L1 loss raised OpenVLA's LIBERO average from 76.5\% to 97.1\% (Section~\ref{sec:posttraining}). \\
Dexterous, bimanual or high-rate control & A flow-matching action expert, as in $\pi_0$~\citep{pi0} & Smooth action chunks at up to 50\,Hz; the current default in frontier systems (Table~\ref{tab:actions}). \\
Simulation benchmark with few demonstrations & VLA-Adapter~\citep{vlaadapter} or VLA-0~\citep{vla0} & Strong LIBERO results with no robot pretraining, but LIBERO tests far less than real-world use (Section~\ref{sec:benchmarks}). \\
Humanoid & GR00T~N1 line~\citep{gr00tn1}, or a VLA over a learned whole-body controller & Open weights and humanoid training data; most humanoid VLAs rely on a separately trained whole-body controller (Section~\ref{sec:humanoid}). \\
Real-robot navigation with on-board compute & Visual navigation foundation models (ViNT, NoMaD, OmniVLA) & Trained on real robot data, output waypoints and target edge deployment (Section~\ref{sec:navigation}). \\
Language-instructed navigation & Video VLMs (NaVid, StreamVLN, NavFoM) & Built for vision-and-language navigation and evaluated on R2R-CE and RxR-CE; their 7--8B models usually run on a remote GPU (Table~\ref{tab:deploy}). \\
Tasks that need minutes of recall & A hierarchical design with memory, as in MemER~\citep{memer} & Keyframe memory kept outside the fast control loop (Section~\ref{sec:inputs}). \\
\bottomrule
\end{tabularx}
\end{table}

\subsection{Model}
\begin{itemize}
  \item \textbf{Inputs and memory.} Start with current camera images, language and robot state. Add history only when the task needs it: long histories cost latency and can teach the policy cues that do not cause the correct action. When memory is needed, compress it, as with keyframes (MemER) or a slow-fast context (StreamVLN), rather than feeding raw history.
  \item \textbf{Architecture.} Start monolithic. Choose a hierarchy when you need interpretable subtasks, live user corrections, long memory or cheap off-domain training data for the high level (Hi Robot, MemER, HAMSTER). Choose a dual system when a large VLM must coexist with fast control (GR00T~N1).
  \item \textbf{Action output.} Predict action chunks rather than single steps. Use flow matching or diffusion for dexterous, high-rate or multimodal behaviour; continuous regression for the fastest inference on narrow tasks; compressed tokens (FAST) to keep a purely autoregressive model; and latent actions to learn from action-free video. When adding a new action expert to a pretrained VLM, protect the VLM with knowledge insulation.
  \item \textbf{Intermediate predictions.} Add reasoning or subtask prediction for long-horizon tasks and open-ended instructions; ECoT gained 28 points over OpenVLA without additional robot data. Predicting future observations is promising but adds cost; consider it when you have plenty of action-free video (VLA-JEPA) or need subgoals for planning ($\pi_{0.7}$).
\end{itemize}

\subsection{Training}
\begin{itemize}
  \item \textbf{Pretraining data.} Favour diversity over volume, as RT-1's ablations first showed. Data from other robots and environments mattered more than web data for $\pi_{0.5}$ in unseen homes, whereas web data mainly helped with unseen objects. When mixing robots, tell the model which robot each example comes from (soft prompts, embodiment prompts or identifier tokens). Keep every episode, including failures: advantage and metadata conditioning can now turn them into useful training data.
  \item \textbf{Post-training.} Several works adapt to a new task with about 50 demonstrations (ACT, TinyVLA, MemER). Use the OpenVLA-OFT recipe for supervised fine-tuning. Move to reinforcement learning from deployment (RECAP, online RL) only when you can run many autonomous trials safely.
\end{itemize}

\subsection{Deployment}
\begin{itemize}
  \item \textbf{Real-time execution.} Measure the latency per action chunk on your own hardware first and compare it with your control period; a 7B autoregressive VLA runs at only a few hertz on one RTX~4090 (Table~\ref{tab:deploy}). If inference is slower than control, execute chunks asynchronously (SmolVLA), add Real-Time Chunking, which needs no retraining, or split a remote VLA from an on-robot adapter when network delays are large (AsyncVLA). Run the levels of a hierarchy at different rates.
\end{itemize}

\subsection{Evaluation and reporting}
\begin{itemize}
  \item Do not rely on LIBERO alone: it is near saturation and hides brittleness. Add a perturbation suite (LIBERO-PRO or LIBERO-Plus) and, where possible, SimplerEnv, whose scores were validated against real robots. For navigation, report R2R-CE and RxR-CE together with real-robot trials.
  \item Compare against a controlled baseline trained with the same data and recipe, as SimVLA advocates, so that a gain can be attributed to the change being studied.
  \item Report the robot episodes and hours in each training stage, the share of non-robot data, the number of embodiments, the parameter count at inference and the latency per chunk on named hardware (Section~\ref{sec:challenges}).
\end{itemize}

\clearpage
\section{Glossary}
\label{app:glossary}

Table~\ref{tab:glossary} explains terms used in this survey that are recent or that come from outside a reader's own field, such as machine-learning terms for readers from classical robotics, robot-data terms for readers from reinforcement learning, and robotics terms for readers from computer vision or language modelling. Terms are grouped by life-cycle stage, as in Section~\ref{sec:taxonomy}.

\definecolor{glossblue}{RGB}{31,78,121}
\newcommand{\glossgroup}[1]{\multicolumn{3}{l}{\rule{0pt}{3ex}\textcolor{glossblue}{\textbf{#1}}} \\[0.3ex]}
{\small
\begin{longtable}{L{3.3cm}L{10.6cm}L{1.3cm}}
\caption{Glossary of terms used in this survey.}\label{tab:glossary}\\
\toprule
Term & Meaning & Section \\
\midrule
\endfirsthead
\toprule
Term & Meaning & Section \\
\midrule
\endhead
\bottomrule
\endfoot
\glossgroup{Model}
Policy & The function that maps what the robot observes to the action it takes; a VLA is one kind of policy. & \ref{sec:background} \\
Vision-language model (VLM) & A model pretrained on web images and text that answers questions about images. VLAs start from one and add robot actions as an output. & \ref{sec:background} \\
Vision-language-action model (VLA) & A VLM adapted to output robot actions from camera images and a language instruction. & \ref{sec:background} \\
Backbone & The pretrained network a VLA is built on, usually the VLM. & \ref{sec:arch} \\
Proprioception & The robot's sensing of its own state, such as joint angles and gripper opening. & \ref{sec:background} \\
Action chunk & A short sequence of future actions predicted in one pass and executed before the model is queried again. & \ref{sec:background} \\
End-effector vs.\ joint-space actions & Actions given as the gripper's target pose, or as targets for each joint. & \ref{sec:manipulation} \\
Relative (delta) actions & Actions expressed as changes from the current state rather than as absolute targets. & \ref{sec:manipulation} \\
Action token & An action written as discrete symbols that a language model can output, for example by binning each action dimension (RT-2, OpenVLA). & \ref{sec:actionout} \\
Autoregressive decoding & Generating output one token at a time, each conditioned on the previous ones. & \ref{sec:actionout} \\
FAST & An action tokeniser that compresses a chunk with the discrete cosine transform before turning it into tokens. & \ref{sec:actionout} \\
Action head / action expert & The part of a VLA that turns the backbone's features into actions. An action expert is a separate set of transformer weights that attends to the VLM's tokens ($\pi_0$). & \ref{sec:actionout} \\
Diffusion policy & A policy that generates actions by starting from noise and removing it step by step, which lets it represent several valid ways to do a task. & \ref{sec:background} \\
Flow matching & A relative of diffusion that learns a velocity field carrying noise to actions, usually in fewer steps. & \ref{sec:actionout} \\
Diffusion transformer (DiT) & A transformer used as the denoising network of a diffusion or flow-matching head. & \ref{sec:manipulation} \\
Monolithic model & One network that maps observations to actions (RT-2, OpenVLA, $\pi_0$). & \ref{sec:arch} \\
Dual-system model & A slow VLM (``System~2'') coupled to a fast action module (``System~1'') through latent features and trained jointly (GR00T~N1). & \ref{sec:arch} \\
Hierarchical model & A high-level model emits an interpretable intermediate, such as a subtask or path, that a separately trained low-level policy executes. & \ref{sec:arch} \\
Key--value (KV) cache & Stored attention states from earlier inputs that a transformer reuses instead of recomputing, which bounds the cost of long histories. & \ref{sec:inputs} \\
\midrule
\glossgroup{Intermediate predictions}
Embodied chain-of-thought & Reasoning written out before acting, such as plans, subtasks or object locations (ECoT). & \ref{sec:intermediate} \\
World model & A model that predicts how the environment will change as the agent acts. & \ref{sec:wam} \\
World-action model (WAM) & A model that predicts future observations and actions together, usually built on a video generator. & \ref{sec:wam} \\
Latent actions & Action-like codes learned from video without action labels, used for pretraining and later mapped to real actions by a small decoder. & \ref{sec:actionout} \\
JEPA & Joint-embedding predictive architecture: predicts the embedding of a future or masked input rather than its pixels. & \ref{sec:wam} \\
\midrule
\glossgroup{Training}
Demonstration & An example of a task performed by a person, usually through teleoperation, from which the policy learns by imitation. & \ref{sec:background} \\
Teleoperation & Collecting demonstrations by a person controlling the robot directly, the main source of robot training data. & \ref{sec:data} \\
Episode & One recorded attempt at a task, from start to end. & \ref{sec:data} \\
Trajectory & The sequence of observations and actions recorded during an episode. & \ref{sec:data} \\
Rollout & One run of a policy on a robot or in simulation; rollouts can be recorded and used for further training. & \ref{sec:posttraining} \\
Embodiment & A particular robot body, with its own sensors and actuators. & \ref{sec:data} \\
Cross-embodiment training & Training one policy on data from several robot types. & \ref{sec:manipulation} \\
Open X-Embodiment (OXE) & A pooled dataset of about one million episodes from 22 robot types, the common pretraining set for open VLAs. & \ref{sec:data} \\
Egocentric human video & Video recorded from a person's viewpoint, such as head- or wrist-mounted cameras, used as a cheap substitute for robot data. & \ref{sec:manipulation} \\
Neural trajectories & Robot training data generated by a video model rather than recorded on a robot (GR00T). & \ref{sec:data} \\
Co-training & Training on robot actions mixed with other data, usually web visual question answering, to keep the VLM's general knowledge. & \ref{sec:data} \\
Post-training & Any training after pretraining that adapts a VLA to a target robot or task: fine-tuning, reinforcement learning or learning from deployment. & \ref{sec:posttraining} \\
LoRA & Low-rank adaptation: fine-tuning a small set of added weights while the original ones stay frozen. & \ref{sec:manipulation} \\
Knowledge insulation & Stopping gradients from a new action expert flowing into the VLM, so that training the expert does not erode the VLM's knowledge. & \ref{sec:actionout} \\
Advantage conditioning & Telling the policy how much better than average each action or episode was, so that failed or autonomous episodes can be used for training ($\pi^*_{0.6}$). & \ref{sec:data} \\
\midrule
\glossgroup{Deployment}
Inference vs.\ control frequency & How often the model is queried, often a few hertz, versus how often the robot receives commands, often tens to hundreds of hertz. Action chunks bridge the two. & \ref{sec:realtime} \\
Latency vs.\ throughput & The delay between an observation and the resulting action, versus the number of actions produced per second. Action chunking can raise throughput without reducing latency. & \ref{sec:realtime} \\
Asynchronous inference & Computing the next chunk while the current one executes, so the robot does not pause. & \ref{sec:realtime} \\
Real-time chunking & Asynchronous inference that keeps fixed the actions about to execute and plans the rest of the next chunk to join them smoothly. & \ref{sec:realtime} \\
Quantisation & Storing weights at lower numerical precision to cut memory and speed up inference. & \ref{sec:manipulation} \\
Edge deployment & Running the model on the robot's own computer rather than on a remote server. & \ref{sec:realtime} \\
\midrule
\glossgroup{Tasks and evaluation}
VLN and VLN-CE & Vision-and-language navigation: following spoken or written route instructions. VLN-CE is the version in continuous environments, with low-level moves instead of jumps between graph nodes. & \ref{sec:navigation} \\
Bimanual manipulation & Tasks that use two arms together, such as folding laundry. & \ref{sec:manipulation} \\
Dexterous manipulation & Fine control of multi-fingered hands or precise grippers, as in tool use or in-hand rotation. & \ref{sec:manipulation} \\
Long-horizon task & A task with many steps, such as cleaning a kitchen, where errors early on compound later. & \ref{sec:manipulation} \\
Closed-loop control & Control that uses fresh observations at every step to correct course; open-loop control executes a plan without looking. & \ref{sec:humanoid} \\
Topological graph & A map of places as nodes connected by traversable paths, without exact metric geometry; used by some navigation systems to plan routes. & \ref{sec:navigation} \\
Loco-manipulation & Tasks that combine moving the body and manipulating objects, such as walking to a table and picking up a cup. & \ref{sec:humanoid} \\
Whole-body control & Coordinating all of a humanoid's joints, legs and arms together, while keeping balance. & \ref{sec:humanoid} \\
Zero-shot & Performing a task, object or setting without any training examples of it. & \ref{sec:manipulation} \\
\end{longtable}
}

\end{document}